\documentclass{ecai} 
\usepackage{latexsym}
\usepackage{amssymb}
\usepackage{amsmath}
\usepackage{amsthm}
\usepackage{booktabs}
\usepackage{enumitem}
\usepackage{graphicx}
\usepackage{color}

\usepackage{hyphenat}

\usepackage{pifont}     % \ding
\usepackage{amssymb}    % \checkmark
\usepackage{array}      % 可选：调整行高
\newtheorem{theorem}{Theorem}

\newtheorem{definition}{Definition}

\newcommand{\BibTeX}{B\kern-.05em{\sc i\kern-.025em b}\kern-.08em\TeX}

\usepackage{latexsym}
\usepackage{amssymb}
\usepackage{amsmath}
\usepackage{amsthm}
\usepackage{booktabs}
\usepackage{enumitem}
\usepackage{graphicx}
\usepackage{color}
\renewcommand{\qedsymbol}{}

\usepackage{algorithm}
\usepackage{algpseudocode}
\usepackage{float}           % 如果想用 [H]，需加载
\usepackage{listings}
\usepackage[T1]{fontenc}
\usepackage{array}
\usepackage{booktabs}
\usepackage[utf8]{inputenc}

\usepackage{listings}
\usepackage{xcolor}
\begin{document}

%%%%%%%%%%%%%%%%%%%%%%%%%%%%%%%%%%%%%%%%%%%%%%%%%%%%%%%%%%%%%%%%%%%%%%%%

\begin{frontmatter}

%%% Use this command to specify your submission number.
%%% In doubleblind mode, it will be printed on the first page.

\paperid{6132} 

%%% Use this command to specify the title of your paper.

% \title{Guidelines for Preparing a Paper for the \\
% European Conference on Artificial Intelligence}
\title{From Bits to Boardrooms: A Cutting-Edge Multi-Agent LLM Framework for Business Excellence}

%%% Use this combinations of commands to specify all authors of your 
%%% paper. Use \fnms{} and \snm{} to indicate everyone's first names 
%%% and surname. This will help the publisher with indexing the 
%%% proceedings. Please use a reasonable approximation in case your 
%%% name does not neatly split into "first names" and "surname".
%%% Specifying your ORCID digital identifier is optional. 
%%% Use the \thanks{} command to indicate one or more corresponding 
%%% authors and their email address(es). If so desired, you can specify
%%% author contributions using the \footnote{} command.

% \author[A]{\fnms{First}~\snm{Author}\orcid{....-....-....-....}\thanks{Corresponding Author. Email: somename@university.edu.}\footnote{Equal contribution.}}
% \author[B]{\fnms{Second}~\snm{Author}\orcid{....-....-....-....}\footnotemark}
% \author[B,C]{\fnms{Third}~\snm{Author}\orcid{....-....-....-....}} 

% \address[A]{Short Affiliation of First Author}
% \address[B]{Short Affiliation of Second Author and Third Author}
% \address[C]{Short Alternate Affiliation of Third Author}

\author[A]{\fnms{Zihao}~\snm{Wang}\orcid{0009-0001-7289-3659}\thanks{Email: wald0wzh@gmail.com.}}
\author[A]{\fnms{Junming}~\snm{Zhang}\orcid{0009-0000-1807-644X}\thanks{Corresponding author. Email: 20241807@hhu.edu.cn.}}

\address[A]{College of Artificial Intelligence and Automation, Hohai University}

%%% Use this environment to include an abstract of your paper.

\begin{abstract}
Large Language Models (LLMs) have shown promising potential in business applications, particularly in enterprise decision support and strategic planning, yet current approaches often struggle to reconcile intricate operational analyses with overarching strategic goals across diverse market environments, leading to fragmented workflows and reduced collaboration across organizational levels. This paper introduces \textbf{BusiAgent}, a novel multi-agent framework leveraging LLMs for advanced decision-making in complex corporate environments. BusiAgent integrates three core innovations: an extended Continuous Time Markov Decision Process (CTMDP) for dynamic agent modeling, a generalized entropy measure to optimize collaborative efficiency, and a multi-level Stackelberg game to handle hierarchical decision processes. Additionally, contextual Thompson sampling is employed for prompt optimization, supported by a comprehensive quality assurance system to mitigate errors. Extensive empirical evaluations across diverse business scenarios validate BusiAgent’s efficacy, demonstrating its capacity to generate coherent, client-focused solutions that smoothly integrate granular insights with high-level strategy, significantly outperforming established approaches in both solution quality and user satisfaction. By fusing cutting-edge AI technologies with deep business insights, BusiAgent marks a substantial step forward in AI-driven enterprise decision-making, empowering organizations to navigate complex business landscapes more effectively.
\end{abstract}

\end{frontmatter}

%%%%%%%%%%%%%%%%%%%%%%%%%%%%%%%%%%%%%%%%%%%%%%%%%%%%%%%%%%%%%%%%%%%%%%%%
\section{Introduction}

\begin{table*}[ht]
\renewcommand{\arraystretch}{0.85}
\small
\centering
\vspace{-3.5em}
\caption{Comparison of multi-agent systems across different capabilities.}
\label{tab:comparison}
\begin{tabular}{lcccc}
\toprule
\textbf{System} & \textbf{Task Delegation} & \textbf{Reporting Work} & \textbf{Prompt Optimization} & \textbf{LLMs Tools} \\
\midrule
Multi-Agent Debate & \ding{55} & \ding{55} & \ding{55} & \ding{55} \\
BabyAGI & \ding{55} & \checkmark & \ding{55} & \ding{55} \\
Camel & \ding{55} & \ding{55} & \checkmark & \ding{55} \\
Chatdev & \checkmark & \ding{55} & \ding{55} & \ding{55} \\
MetaGPT & \checkmark & \ding{55} & \ding{55} & \ding{55} \\
Generative Agents & \ding{55} & \ding{55} & \ding{55} & \ding{55} \\
METAAGENTS & \ding{55} & \ding{55} & \ding{55} & \ding{55} \\
\textbf{BusiAgent (Ours)} & \checkmark & \checkmark & \checkmark & \checkmark \\
\bottomrule
\end{tabular}
\end{table*}

In the era of digital transformation, modern enterprises face a central challenge: converting extensive operational data into impactful boardroom decisions across diverse contexts. This journey necessitates balancing fine-grained analytics with macro-level strategy, requiring organizations to orchestrate complex information flows across multiple decision-making layers. The exponential growth of business data—from customer interactions to internal operations—has created unprecedented opportunities alongside significant coordination challenges for executive teams.

While Large Language Models (LLMs) have demonstrated notable potential in business applications~\cite{hong2023metagpt,tang2025autoagent,jimenez2025multi}, prevailing methods fragment workflows and underutilize synergy across organizational tiers~\cite{park2023generative,zhuge2023mindstorms,cai2023large,wang2023unleashing}. These approaches treat business problems as isolated domains rather than interconnected ecosystems, limiting their effectiveness where cross-functional alignment is critical. Consequently, they fail to align operational insights with strategic outcomes, undermining cross-departmental collaboration and creating decision silos that hinder organizational agility.

\begin{figure}[t]
\centering
\includegraphics[width=0.8\columnwidth]{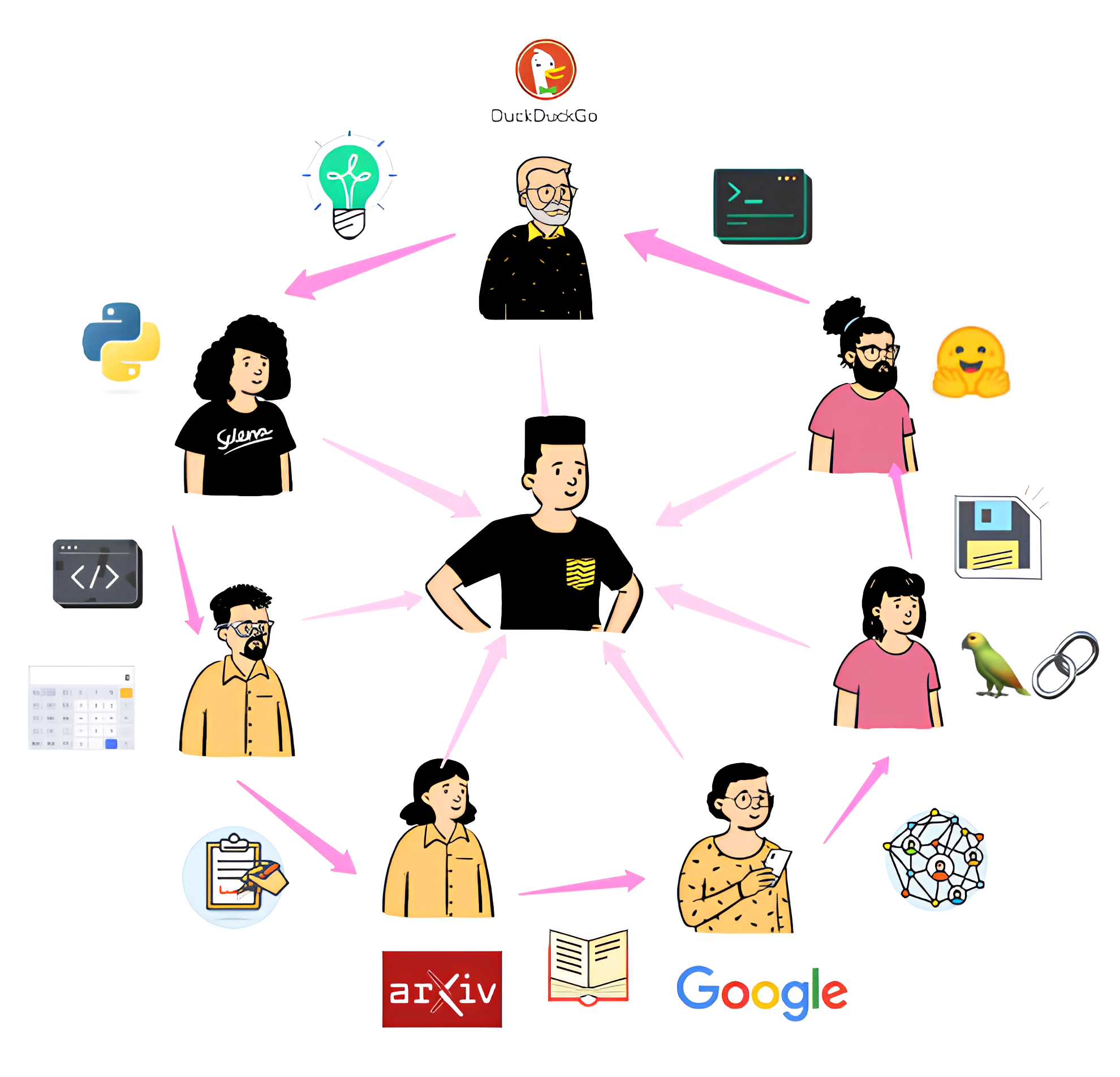}
\vspace{-1em}
\caption{Client-centric Multi-agent System}
\label{fig:client-system}
\vspace{-1em}
\end{figure}

This misalignment between operational granularity and strategic coherence manifests in critical challenges: (1) information asymmetry across management layers, (2) delayed decision-making from fragmented analysis, (3) disconnect between business objectives and daily operations. McKinsey research~\cite{rossetto2021conservation} indicates that companies excelling at integrating data-driven insights across organizational levels achieve 5-6\% higher productivity and profitability.

Through extensive practice, structured workflows~\cite{rossetto2021conservation} have emerged to systematically connect operational data with strategic initiatives, especially for multinational enterprises. These workflows break complex processes into well-defined phases, establishing transparent task sequences while enabling automation. However, implementing such frameworks requires substantial manual coordination and cross-domain expertise—constraints limiting scalability in evolving markets.

To address these challenges, we introduce \textbf{BusiAgent}, a GPT-based multi-agent framework harnessing organizational workflows to fulfill diverse client needs~\cite{rossetto2021conservation}. As depicted in Figure~\ref{fig:client-system}, BusiAgent orchestrates role-specific agents (CEO, CTO, CFO, Marketing Manager) collaborating through structured procedures, supplemented by specialized tools for coding, search, and analysis. Unlike previous approaches focusing on singular business aspects, BusiAgent mimics organizational dynamics through hierarchical decision-making enabling horizontal collaboration and vertical coordination.

The framework incorporates a Continuous Time Markov Decision Process (CTMDP) for dynamic agent modeling, preserving structured workflows for task allocation, role responsibilities, and tool utilization. This mathematical foundation enables BusiAgent to balance exploration and exploitation across business contexts. The architecture delivers cohesive solutions by aligning high-level strategy with operational execution, boosting client satisfaction and adapting to evolving demands.

\noindent\textbf{Contributions.} We summarize our contributions as follows:
\begin{itemize}
\item \textbf{BusiAgent:} A novel LLM-based multi-agent framework that transforms operational data into strategic insights, integrating extended CTMDP, information theory, and multi-level Stackelberg games to enable data-driven boardroom decisions. Our approach demonstrates significant improvements in solution quality (+122\% in problem analysis, +284\% in task assignment) compared to state-of-the-art baselines, \textit{which aligns with McKinsey’s latest projection that multi-agent AI could raise enterprise productivity by roughly 1.5 percentage points per year}~\cite{mckinsey2025ai}.

\item \textbf{Cognitive Enhancement:} An adaptive system combining contextual Thompson sampling for prompt optimization and comprehensive quality assurance mechanisms, ensuring reliable translation of operational data into strategic recommendations. This dual approach mitigates hallucination risks while maximizing contextual relevance, achieving a 4.30/5.0 user satisfaction rating across diverse business scenarios.

% \item \textbf{AI-Human Parallel:} An innovative approach aligning AI decision-making processes with corporate management structures, demonstrating superior performance in bridging operational analysis and strategic planning. Our extensive evaluation with 100 domain experts validates BusiAgent's effectiveness in generating solutions that seamlessly integrate granular insights with strategic vision.
\item \textbf{AI-Human Parallel:} An innovative approach aligning AI decision-making processes with corporate management structures, demonstrating superior performance in bridging operational analysis and strategic planning. Our extensive evaluation with 100 domain experts validates BusiAgent's effectiveness in generating solutions that seamlessly integrate granular insights with strategic vision, mirroring human executive decision-making.
\end{itemize}

\begin{figure*}[!h]
\centering
\vspace{-3em}
\includegraphics[width=0.96\linewidth]{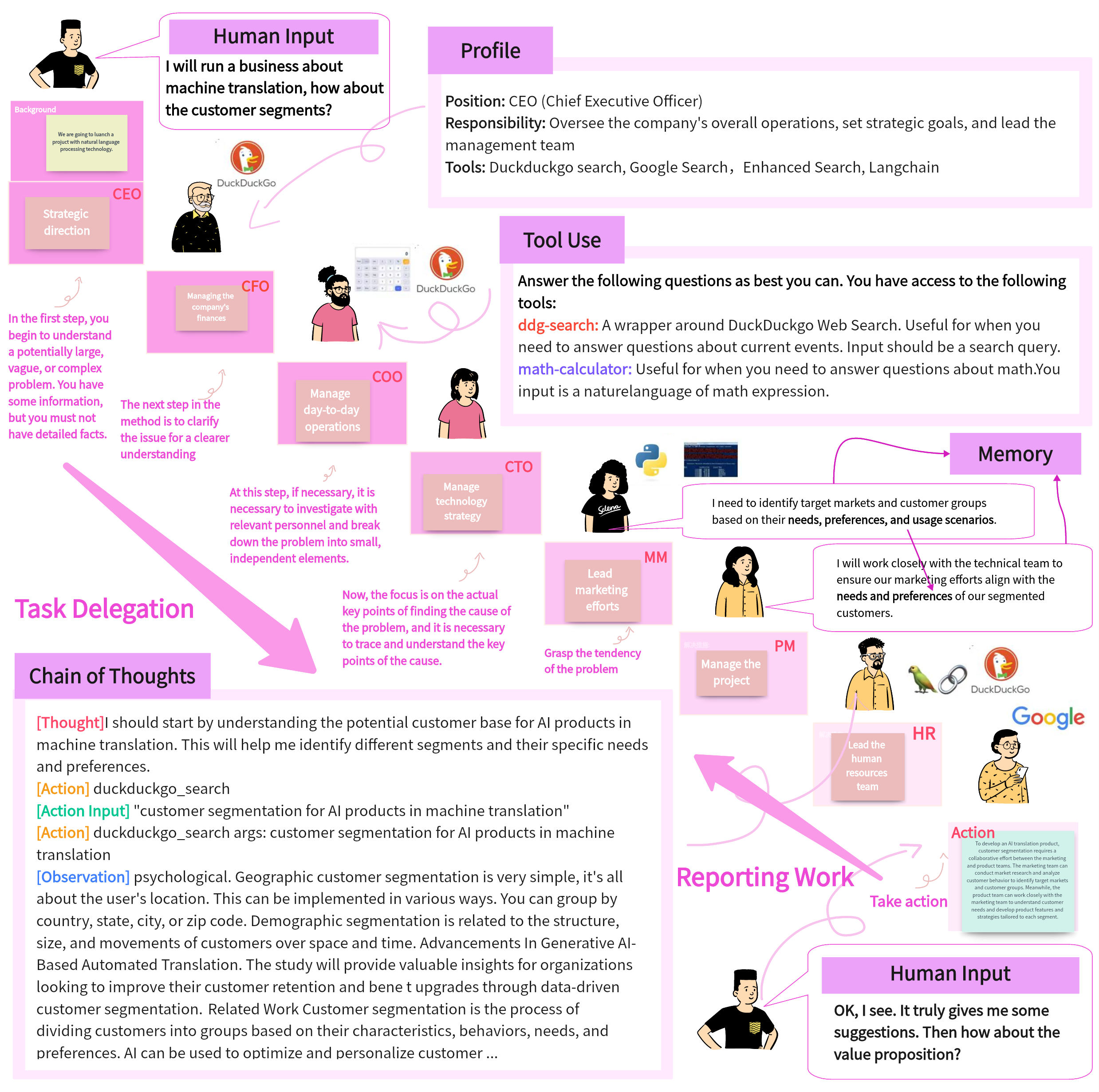}
\caption{BusiAgent: A Client-Centric Business Framework. See Appendix F for a detailed explanation.}
\label{framework}
%\vspace{-3pt}
\end{figure*}
\vspace{-2em}

\section{Related Work}
\subsection{LLM Based Agent}
Large language models have emerged as core components of AI agents due to their capabilities~\cite{openai2023gpt4,claude3,brown2020language,gemini15,gpt4o2024,ouyang2022training,wei2022emergent}. Research has focused on extending LLMs beyond language processing to include multimodal perception, improved perceptual skills, and tool usage techniques~\cite{autogenstudio2024,llmagentSurvey2025,lu2024chameleon,nakano2021webgpt,qin2023tool,schick2023toolformer,yao2022react}, signifying a substantial shift toward more adaptable AI agents.

Based on LLMs, various agents have been developed, each with unique capabilities. AutoGPT \cite{Significant_Gravitas_AutoGPT} automates complex tasks into subtasks with web integration. WorkGPT \cite{WorkGPT} engages in dialogues with LLMs for task execution. GPT-Engineer \cite{gpt-engineer} and SmolModels \cite{SmolModels} focus on automating code generation. ChatGPT-Code Interpreter \cite{openai_2023_CI} combines conversational AI with a code interpreter, and LangChain Agents \cite{xagent2024,langchain_2023_agents,autoagent2025,agentverse2023} offers different types of agents for LLM-based applications. It provides a natural-language API with curated tools, following a single-agent paradigm.

\subsection{Multi-Agent Systems and Scenarios}
Recent multi-agent systems have leveraged multiple LLMs for enhanced reasoning through inter-agent discussions \cite{du2023improving,masllm2025,li2023camel,talkhier2025,liang2023encouraging,multiagentbench2025,wu2023autogen,sirius2025}. Notable examples include BabyAGI for task management, CAMEL for role-playing task completion, MetaGPT for software development, and AutoGen as an open-source framework for diverse applications \cite{hong2023metagpt,autoagent2025,agentverse2023}.

As illustrated in Table~\ref{tab:comparison}, we provide a systematic comparison of existing multi-agent systems across four critical capabilities: task delegation, reporting work, prompt optimization, and LLM tools integration. While systems like ChatDev and MetaGPT excel in task delegation, and others like BabyAGI in reporting work or Camel in prompt optimization, none comprehensively addresses all four dimensions. BusiAgent uniquely integrates all these capabilities, enabling more cohesive workflows that bridge operational details with strategic insights.

% The applications span various domains: MetaAgent \cite{li2023metaagents} in job fair simulations, MetaGPT and ChatDev \cite{qian2023communicative} in software development, and AutoGen across mathematics, coding, and operations research. While current implementations focus on specific scenarios, our work provides more comprehensive simulations across multiple business sectors.
The applications span various domains: MetaAgent \cite{li2023metaagents} in job fair simulations, MetaGPT \cite{hong2023metagpt} and ChatDev \cite{qian2023communicative} in software development, and AutoGen across mathematics, coding, and operations research \cite{multiagentbench2025}. While current implementations focus on specific scenarios, this research significantly extends prior work by providing more comprehensive and diverse simulations across multiple business sectors, demonstrating unprecedented versatility and practical utility.

\section{BusiAgent: Multi-Agent Business Workflow Framework}

Business process optimization requires coordinated decision-making across organizational levels while maintaining operational efficiency. Previous approaches often struggle with integrating strategic planning and operational execution. We propose BusiAgent, a novel multi-agent framework that enhances business workflows through specialized AI agents, making three key technical contributions:
\begin{enumerate}
\item A \textbf{role-based agent system} using extended Continuous Time Markov Decision Process (CTMDP),
\item A \textbf{collaborative decision-making mechanism} combining entropy-based discussions and multi-level Stackelberg games,
\item An \textbf{enhanced decision support system} with contextual Thompson sampling and robust quality assurance.
\end{enumerate}

Figure~\ref{framework} illustrates the overall architecture.

\subsection{Framework Overview}

BusiAgent unifies strategic planning and operational execution through five integrated components:

\noindent
(1) \textbf{Role-based agent system} uses extended CTMDPs to optimize decisions among specialized roles (CEO, CFO, CTO, etc.).

\noindent
(2) \textbf{Collaborative decision-making mechanism} combines peer-level entropy-based brainstorming with hierarchical Stackelberg coordination.

\begin{figure}[t]
\centering
\vspace{-1em}
\includegraphics[width=\columnwidth]{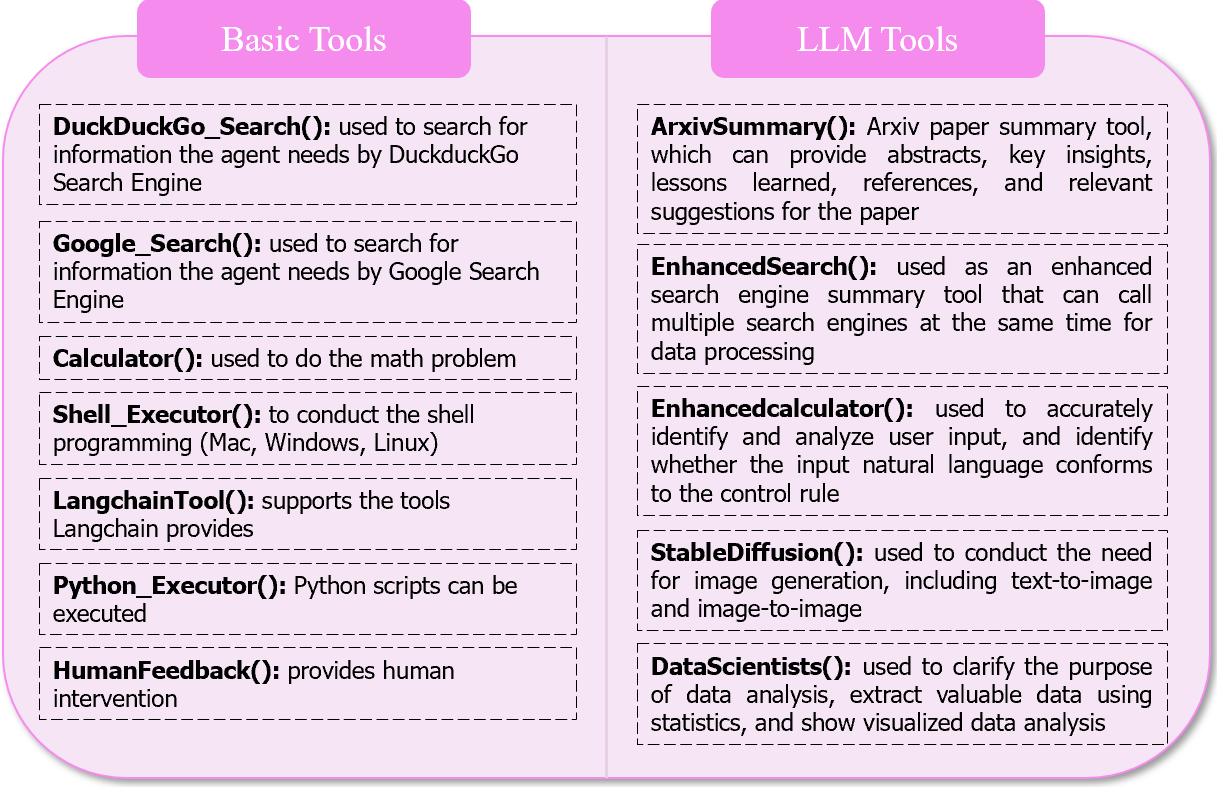}
\vspace{-1em}
\caption{Examples of Tools in BusiAgent. See Appendix F for a detailed explanation.}
\label{fig:tools}
\end{figure}

\noindent
(3) \textbf{Tool integration system} extends action spaces with specialized business tools (Figure~\ref{fig:tools}).

\noindent
(4) \textbf{Advanced prompt optimization} leverages contextual Thompson sampling to refine LLM queries dynamically.

\noindent
(5) \textbf{Quality assurance system} merges short-/long-term memory and a knowledge base to ensure correctness.

\begin{figure}[t]
\centering
\vspace{-1em}
\includegraphics[width=\columnwidth]{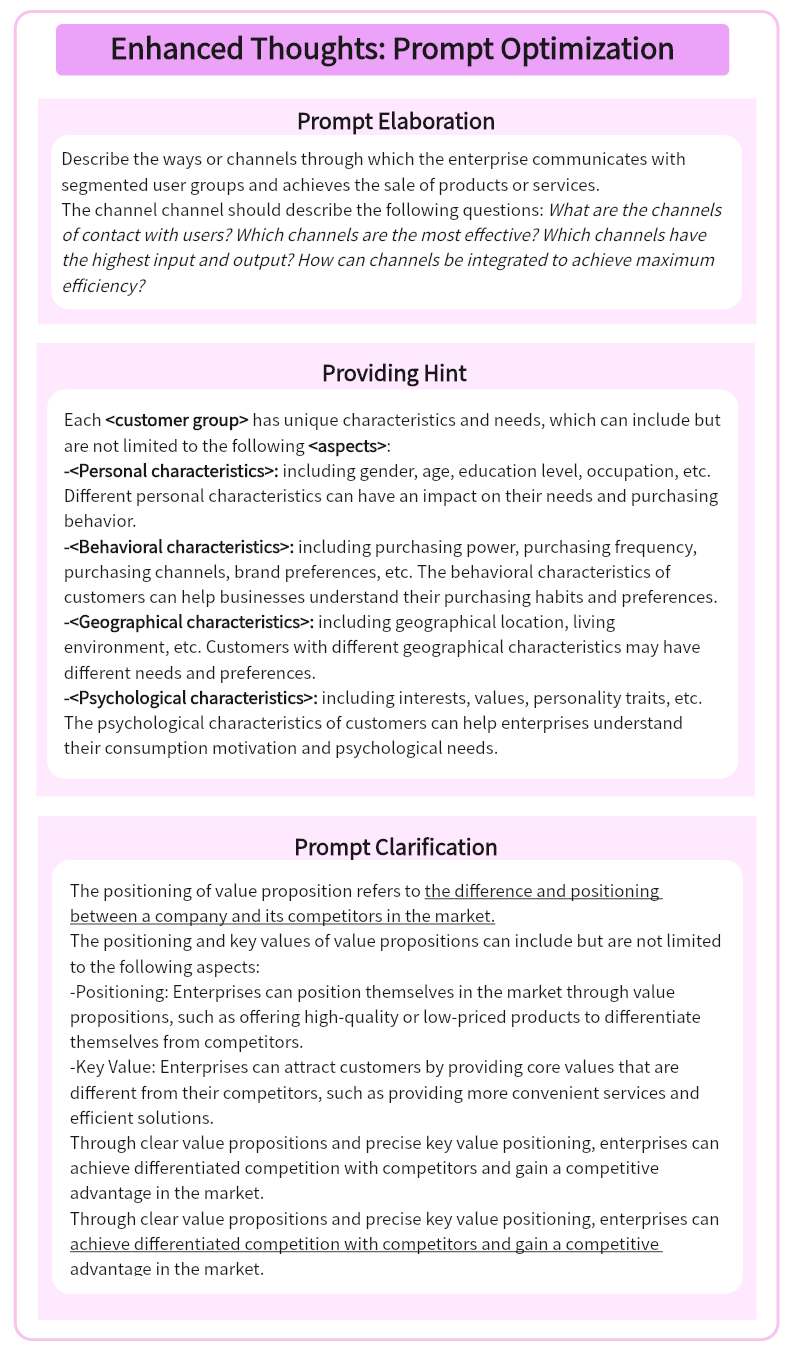}
\vspace{-1em}
\caption{Enhanced Thoughts: Prompt Optimization. See Appendix F for a detailed explanation.}
\label{fig:prompt-optimization}
\end{figure}

\begin{figure}[t]
\centering
\vspace{-1em}
\includegraphics[width=\columnwidth]{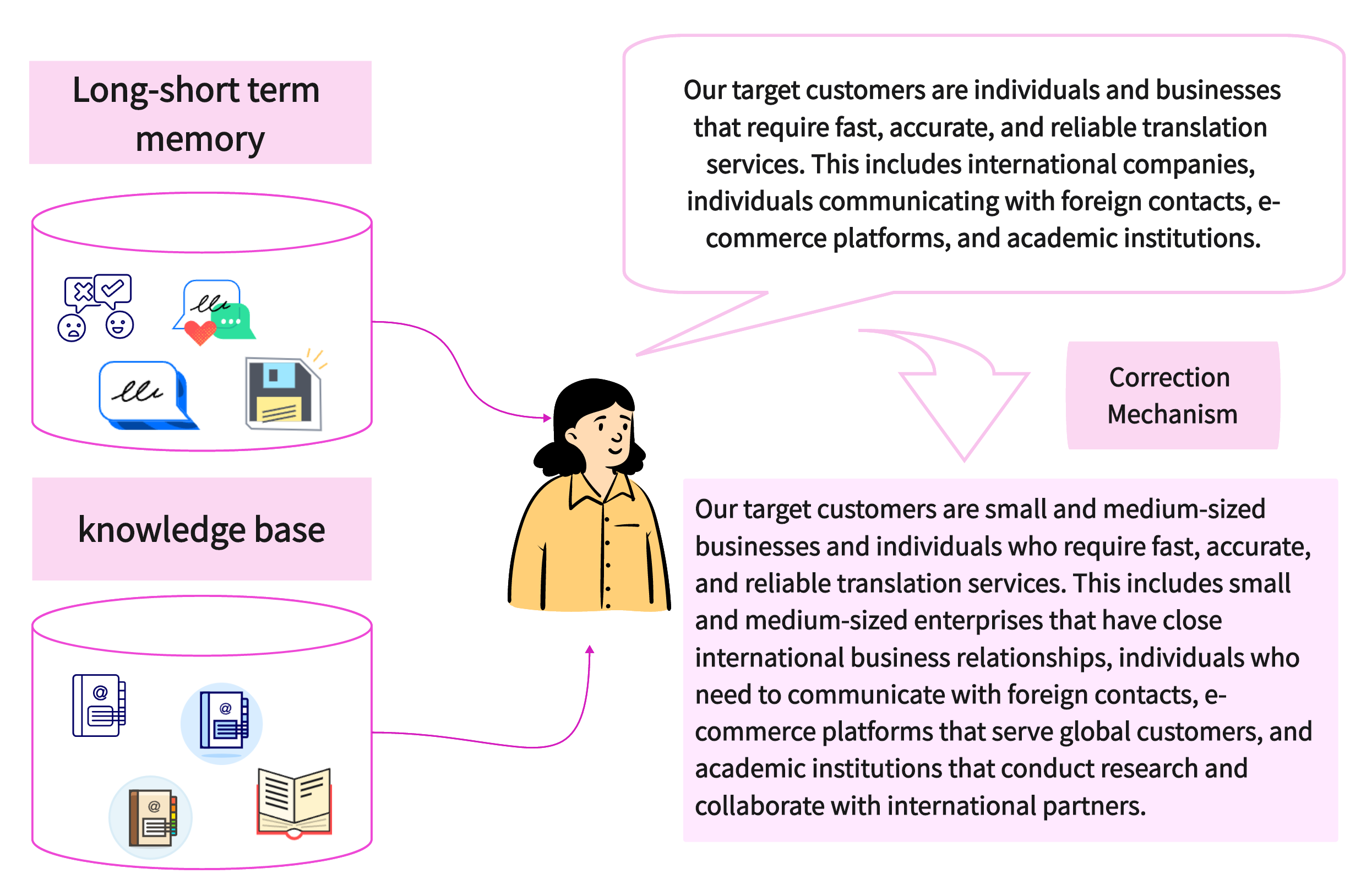}
\vspace{-1em}
\caption{Quality Assurance Mechanism: LTM + STM + Knowledge Base. See Appendix F for a detailed explanation.}
\label{fig:correction}
%\vspace{-8pt}
\end{figure}

\subsection{Role-Based Agent System}
\label{Extended-CTMDP}

Modern organizations rely on domain-specific roles with unique responsibilities. BusiAgent formalizes each role $i$ with an \emph{extended CTMDP}:

\begin{definition}[Extended CTMDP for BusiAgent Roles]
\label{def:CTMDP}
For each agent $i$, define $(S_i,A_i,q_i,r_i,\gamma_i,\omega_i)$:
\begin{itemize}
\item $S_i$: state space (role-specific business contexts),
\item $A_i$: action space (both strategic and operational),
\item $q_i: S_i\times A_i\times S_i\to\mathbb{R}^+$: transition rate function,
\item $r_i: S_i\times A_i\to\mathbb{R}$: reward rate function,
\item $\gamma_i\in(0,1)$: continuous-time discount factor,
\item $\omega_i: S_i\times A_i\to\mathbb{R}^+$: duration or deadline function.
\end{itemize}
\end{definition}

This extension allows explicit modeling of time-sensitive tasks.

\begin{theorem}[Extended Optimality Equation for BusiAgent CTMDP]
\label{thm:ctmdp_main_body}
For agent $i$, the optimal value function $V_i^*(s)$ satisfies:
\begin{equation}
\label{eq:ctmdp_main}
\begin{aligned}
V_i^*(s) \;=\;\max_{a\in A_i}\Bigl[
&\,r_i(s,a)\,\omega_i(s,a)\;+\;\\
&\;\;\int_{0}^{\omega_i(s,a)}e^{-\gamma_i t}\sum_{s'\in S_i}q_i(s,a,s')\,V_i^*(s')\,dt
\Bigr].
\end{aligned}
\end{equation}
\textit{(See expanded proof in Appendix~C.1.)}
\end{theorem}

\subsection{Collaborative Decision Making}
\label{Stackelberg-game}

Horizontal (peer-level) collaboration is achieved via an entropy-based brainstorming mechanism, while vertical (hierarchical) collaboration is handled through a multi-level Stackelberg game.

\paragraph{Horizontal Collaboration.} We define a generalized entropy:

\begin{definition}[Generalized Entropy of Brainstorming]
\label{def:entropy}
For discussion outcome $X$,
\begin{equation}
\label{eq:entropy_brainstorm}
H_{\alpha}(X)=\frac{1}{1-\alpha}\,\log \sum_{x\in\mathcal{X}}p(x)^\alpha q(x)^{1-\alpha},
\end{equation}
where $\alpha>0,\alpha\neq 1$, $p(x)$ is the posterior, and $q(x)$ is the prior distribution.
\end{definition}

\begin{theorem}[Generalized Efficiency of Brainstorming]
\label{thm:entropy_main_body}
If the Rényi divergence $D_{\alpha}(p\|q)\ge \epsilon>0$, BusiAgent’s brainstorming reduces the expected solution time by at least a factor $2^\epsilon$.  
\textit{(Appendix~C.2 contains the extended proof.)}
\end{theorem}

\paragraph{Vertical Coordination.} We apply multi-level Stackelberg games:

\begin{definition}[Multi-Level Stackelberg Game in BusiAgent]
\label{def:stackelberg}
A multi-level Stackelberg game is $(N,L,(S_i),(U_i),(f_l))$, with:
\begin{itemize}
\item $N$: set of agents,
\item $L$: set of hierarchical levels,
\item $S_i$: strategy sets, $U_i$: utility functions,
\item $f_l$: decision function at level $l$.
\end{itemize}
\end{definition}

\begin{theorem}[Equilibrium in Multi-Level BusiAgent Game]
\label{thm:stackelberg_main_body}
Under mild conditions, there exists a unique subgame perfect equilibrium:
\begin{equation}
\label{eq:stackelberg_main}
f_l^*(s)=\arg\max_{a_l\in A_l}U_l\bigl(s,a_l,f_{l+1}^*(s)\bigr).
\end{equation}
\textit{(Appendix~C.3 includes a thorough existence/uniqueness proof.)}
\end{theorem}

\begin{figure*}[t]
\centering
\vspace{-3.5em}
\includegraphics[width=0.9\textwidth]{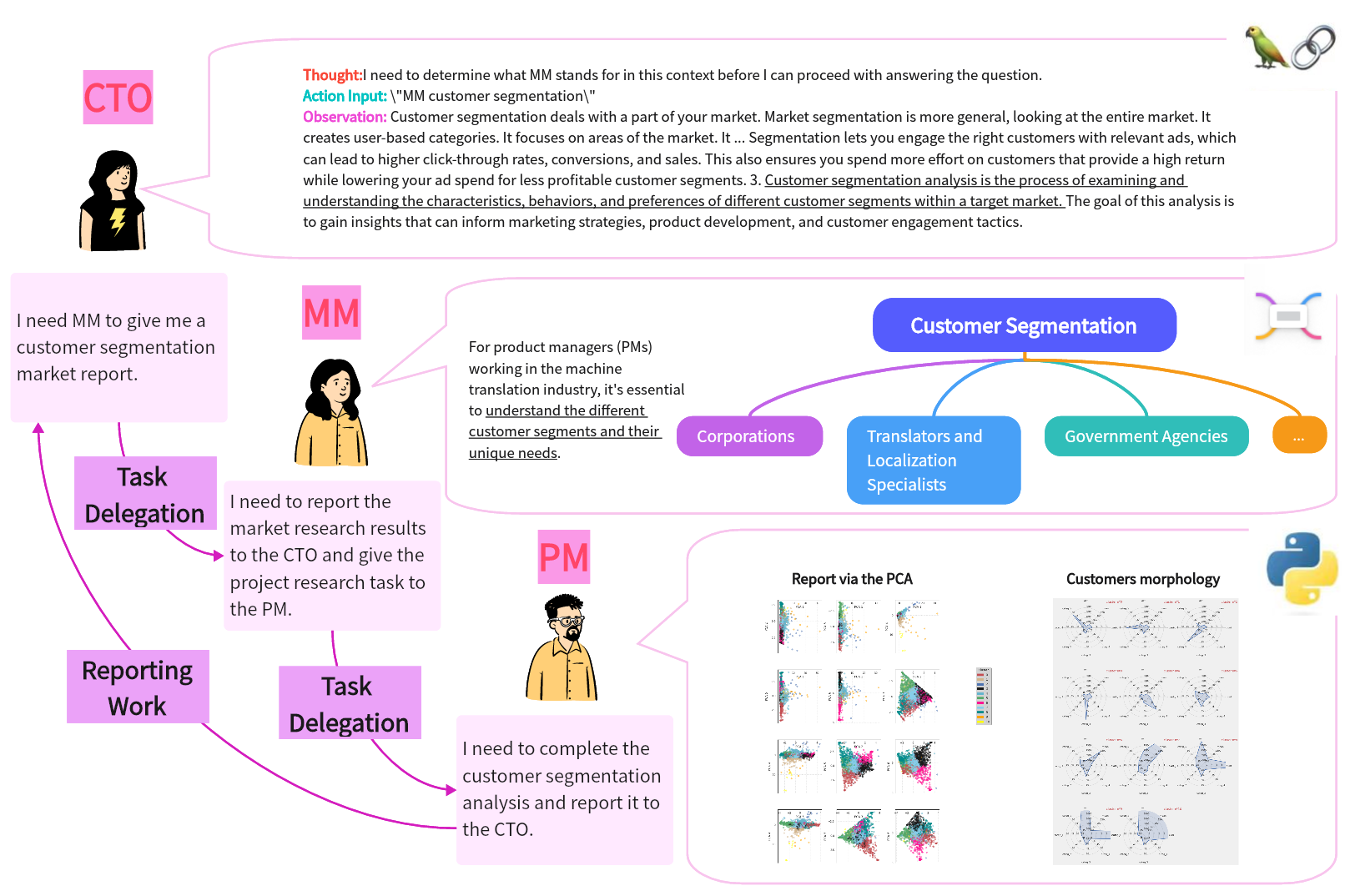}
\caption{Customer Segmentation Analysis: CEO delegates tasks to CTO; CTO coordinates Marketing Manager (MM) and Product Manager (PM). PM invokes Python for data analytics. See Appendix F for a detailed explanation.}
\label{fig:customer-segmentation}
\vspace{-7pt}
\end{figure*}

\vspace{-0.5em}
% \subsection{Enhanced Decision Support}

% BusiAgent’s decision support is strengthened by \emph{tool integration} and \emph{contextual Thompson sampling}.

% \paragraph{Tool Integration.} The action space $A_i$ for each role includes specialized “tools,” as in Figure~\ref{fig:tools}, enabling immediate data retrieval, coding, or analytics.

% \paragraph{Contextual Thompson Sampling.} We use a multi-armed bandit approach with Gaussian process priors for different prompt variants:

% \begin{definition}[Contextual Thompson Sampling for Prompt Optimization]
% \label{def:cts}
% For each variant $k$ and context $x$, define $\mathcal{GP}(\mu_k,K_k)$. At round $t$:
% \begin{enumerate}
% \item Observe context $x_t$,
% \item Sample $\theta_k(t)$ from posterior $\mathcal{GP}(\mu_k,K_k)$,
% \item Pick $k_t=\arg\max_{k}\theta_k(t)(x_t)$,
% \item Observe reward $r_t\in[0,1]$,
% \item Update to $\mathcal{GP}(\mu_k',K_k')$.
% \end{enumerate}
% \end{definition}

% \begin{theorem}[Regret Bound for Contextual Thompson Sampling]
% \label{thm:ts_main_body}
% We have 
% \begin{equation}
% \label{eq:thompson_regret}
% \mathbb{E}[R(T)]\le O\Bigl(\sqrt{d\,K\,T\,\gamma_T\,\log T}\Bigr),
% \end{equation}
% where $d$ is context dimension, $K$ the number of prompt variants, and $\gamma_T$ the maximum information gain.  
% \textit{(See Appendix~C.4 for detailed derivation.)}
% \end{theorem}
\subsection{Enhanced Decision Support}

BusiAgent’s decision support is strengthened by two key components: 
\emph{tool integration} and \emph{contextual Thompson sampling}.

\paragraph{Tool Integration.}
The action space $A_i$ for each role includes specialized “tools” (Figure~\ref{fig:tools}), 
enabling immediate data retrieval, coding, or analytics. By embedding these tools in the CTMDP
action set, each agent (CEO, CFO, etc.) can dynamically decide to call 
(\texttt{DuckDuckGo\_Search()}, \texttt{Python\_Executor()}, etc.) 
to gather information or perform computations relevant to its domain.

\paragraph{Contextual Thompson Sampling.}
We use a multi-armed bandit approach with Gaussian process priors for different prompt variants:

\begin{definition}[Contextual Thompson Sampling for Prompt Optimization]
\label{def:cts}
For each variant $k$ and context $x$, define $\mathcal{GP}(\mu_k,K_k)$. At round $t$:
\begin{enumerate}
\item Observe context $x_t$,
\item Sample $\theta_k(t)$ from posterior $\mathcal{GP}(\mu_k,K_k)$,
\item Pick $k_t=\arg\max_{k}\theta_k(t)(x_t)$,
\item Observe reward $r_t\in[0,1]$,
\item Update to $\mathcal{GP}(\mu_k',K_k')$.
\end{enumerate}
\end{definition}

\begin{theorem}[Regret Bound for Contextual Thompson Sampling]
\label{thm:ts_main_body}
We have 
\begin{equation}
\label{eq:thompson_regret}
\mathbb{E}[R(T)]\;\le\;O\Bigl(\sqrt{\,d\,K\,T\,\gamma_T\,\log T}\Bigr),
\end{equation}
where $d$ is the context dimension, $K$ the number of prompt variants, 
and $\gamma_T$ the maximum information gain.
\textit{(See Appendix~C.4 for a detailed derivation.)}
\end{theorem}

\noindent
\textbf{Figure~\ref{fig:prompt-optimization}} provides a visual illustration of how 
BusiAgent \emph{refines} user prompts throughout this sampling process, breaking each prompt 
into phases such as \emph{Prompt Elaboration}, \emph{Providing Hint}, and \emph{Prompt Clarification}. 
During each iteration, the agent may expand on the user’s original question, highlight relevant 
dimensions (e.g., personal or behavioral characteristics), and clarify core business concepts 
like value proposition. By evaluating the ``reward'' (i.e., the improved clarity and alignment 
of the prompt), the system adaptively converges on more effective instructions. 
This mechanism avoids misunderstandings and ensures that each agent role (CEO, CFO, etc.) 
receives contextually optimized tasks that lead to higher-quality business solutions.

\subsection{Quality Assurance Mechanism}
\label{qa-system}

Finally, BusiAgent integrates a robust QA system (Figure~\ref{fig:correction}), merging short-/long-term memory with a knowledge base to detect and correct potential inconsistencies. This ensures outputs remain aligned with overarching business objectives.

The system combines three key components working in concert. Short-term memory captures immediate conversational context and partial solution states, while long-term memory preserves historical patterns including past agent decisions and domain constraints. These memories enable the system to verify whether new suggestions conflict with established knowledge. Complementing this dual-memory architecture, a curated knowledge base serves as a repository of business guidelines, industry standards, and domain-specific information.

When inconsistencies arise—such as proposals violating budget constraints or contradicting best practices—the correction mechanism triggers automatically. BusiAgent prompts the responsible role to revise recommendations or requests user clarification, repeating until reaching a consistent outcome. This feedback loop prevents agent "drift" and minimizes downstream rework.

In practice, this might manifest when refining target customer segments for a service. If the Marketing Manager proposes new segments based on analytics, the QA system automatically checks previously documented budget limitations and market entry guidelines. Upon detecting conflicts (e.g., expansion costs exceeding approved budgets), the system flags the discrepancy and initiates cross-functional resolution. This multi-layered approach ensures all solutions pass rigorous validation, maintaining alignment between immediate tactical decisions and long-term strategic vision throughout extended interactions.

\section{Experimental Evaluation}
\label{experiment}

\subsection{Implementation Details}

The framework is primarily implemented with \texttt{text-davinci-003} as the LLM, also tested on \texttt{gpt-3.5-turbo-16k-0613}, \texttt{gpt-4o-2024-11-20}, \texttt{Llama-3-70b}, and \texttt{Guanaco-65B-GPTQ} to evaluate generality across architectures.

\subsection{Business Simulation: Serving a Client}
\label{sec:business-sim}

This section demonstrates BusiAgent's practical capabilities through a \emph{customer segmentation} case study for a machine translation startup. Figure~\ref{fig:customer-segmentation} illustrates how the organizational hierarchy processes complex analytics requirements through structured role delegation.

The simulation initiates with the CEO establishing a clear strategic objective—identifying distinct customer segments for targeted service development. This directive flows through the organizational structure following a defined hierarchy:

\begin{itemize}
\item \textbf{CEO $\to$ CTO}: The executive articulates the high-level objective ("customer segmentation for strategic planning") and delegates technical coordination responsibilities.
\item \textbf{CTO $\to$ Marketing Manager}: The technology officer assigns market research components, leveraging specialized expertise in demographic analysis and preference modeling.
\item \textbf{Marketing Manager $\to$ Product Manager}: Following initial data collection, the Product Manager conducts quantitative analytics utilizing advanced clustering methodologies and principal component analysis.
\end{itemize}

The Product Manager executes specialized Python-based analytics through the \texttt{Python\_Executor} tool (Section~3.4), generating visualization outputs including scatter plots and morphological clusters. These results flow upward through the hierarchy—PM $\to$ CTO—for technical validation before reaching the CEO for strategic integration. This orchestrated workflow exemplifies three fundamental aspects of the framework:

\begin{enumerate}[label=(\alph*)]
\item \emph{Task Delegation}: Systematic distribution of responsibilities through the organizational chain.
\item \emph{Reporting Work}: Structured upward flow of analytics for comprehensive quality assurance.
\item \emph{Role-Specific Actions}: Specialized operations performed by appropriate roles, aligned with the extended CTMDP model (Section~3.2).
\end{enumerate}

This structured approach ensures segmentation insights achieve both \emph{comprehensive scope} and \emph{analytical rigor}. By connecting strategic vision with operational execution, BusiAgent effectively minimizes information fragmentation while maintaining decision alignment with organizational objectives and resource constraints.

\subsection{Performance Analysis}

% \noindent\textbf{Dataset.} Use the \emph{AI Company Generation} set with 100 tasks across:
% \begin{itemize}
% \item Problem Analysis ($n=30$),
% \item Task Assignment ($n=30$),
% \item Solution Development ($n=40$).
% \end{itemize}
\noindent\textbf{Dataset.} The evaluation utilized the \emph{AI Company Generation} dataset comprising 100 diverse business tasks spread across three categories: Problem Analysis (30 tasks), Task Assignment (30 tasks), and Solution Development (40 tasks). This balanced distribution enabled comprehensive assessment of BusiAgent's capabilities across the entire business decision-making pipeline, from initial problem framing to final solution implementation.

\noindent\textbf{Human Expert Evaluations.} 100 domain experts provided 941 ratings on solution completeness, coherence, feasibility.

\begin{figure}[t]
\centering
\includegraphics[width=\columnwidth]{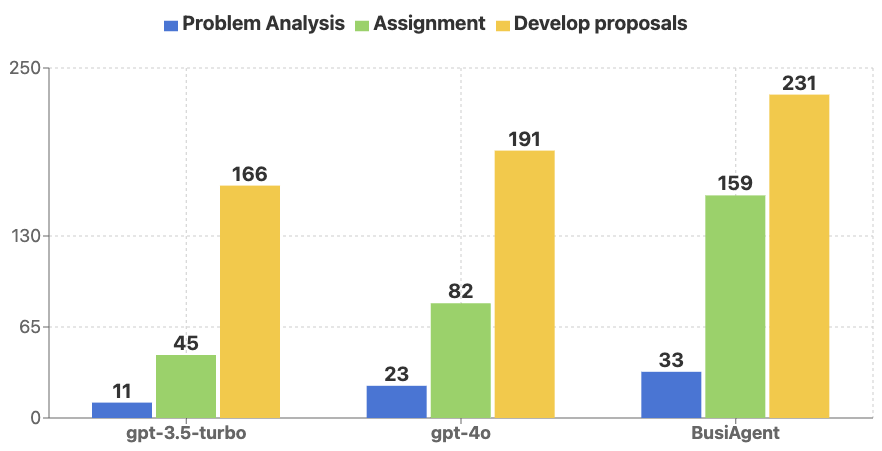}
\vspace{-2em} % -3.5pt
\caption{Expert Voting Results Across Problem-solving Categories and System Variants}
\label{fig:voting-results}
\vspace{-0.5em}
\end{figure}

As shown in Figure~\ref{fig:voting-results}, BusiAgent significantly outperforms baselines (e.g., +122\% in Problem Analysis, +284\% in Task Assignment). Welch’s $t$-tests confirm $p<0.05$.  

% \noindent\textbf{Token Usage.} Table~\ref{tab:token-consumption} shows token consumption under different models. BusiAgent uses more tokens but provides deeper multi-agent reasoning.
\noindent\textbf{Token Usage.} Table~\ref{tab:token-consumption} shows token consumption under different models for each task category. BusiAgent uses more tokens but provides deeper multi-agent reasoning through collaborative role-based interactions and hierarchical processing of business tasks.

\begin{table}[t]
\renewcommand{\arraystretch}{0.85}  % Reduced row height
\setlength{\tabcolsep}{4pt}         % Reduced column spacing
\centering
\caption{Generation Token Consumption Across LLMs and Task Types}
\vspace{-5pt}
\small
\label{tab:token-consumption}
\begin{tabular}{@{}lccccc@{}}  % First column left-aligned, others centered
\toprule
\small
Task & gpt-3.5 & gpt-4o & Llama-3 & Guanaco & BusiAgent \\
\midrule
Analysis       & 171  & 343  & 201  & 200  & \textbf{1148} \\
Assignment     & 65   & 97   & 53   & 50   & \textbf{229}  \\
Development    & 998  & 1264 & 896  & 889  & \textbf{4748} \\
\bottomrule
\end{tabular}
\end{table}

% \noindent\textbf{User Satisfaction.} On a 5-point Likert scale, BusiAgent achieves 4.30 vs. 3.87 (GPT-4o) and 3.55 (GPT-3.5), with experts praising synergy and domain awareness.
\noindent\textbf{User Satisfaction.} On a 5-point Likert scale, BusiAgent achieves 4.30 vs. 3.87 (GPT-4o) and 3.55 (GPT-3.5), with experts praising synergy and domain awareness across diverse business scenarios requiring multi-level organizational coordination.

\subsection{Organizational Dynamics Analysis}

Evaluate dynamic scenarios: role dependencies, workflow adaptation, collaboration patterns, tested via ablations and user surveys. 

\paragraph{Role Dependence.}  
Table~\ref{tab:role-dependence} shows average tokens per role and “dependence.” The CEO has the highest cross-role calls (6). The PM uses the most tokens (984).

\begin{table}[t]
\renewcommand{\arraystretch}{0.75}
\centering
\caption{Token Consumption and Dependence by Role}
\vspace{-3.5pt}
\small
\label{tab:role-dependence}
\begin{tabular}{@{}lccccccc@{}}
\toprule
\small
Role & CEO & CFO & COO & CTO & MM & PM & HR \\
\midrule
Avg Token   & 679 & 522 & 824 & 891 & 711 & \textbf{984} & 463 \\
Dependence  & \textbf{6} & 3 & 3 & 3 & 2 & 1 & 1 \\
\bottomrule
\end{tabular}
\end{table}

Using MovieLens 100K~\cite{harper2015movielens}, we remove roles. Table~\ref{tab:movielens} shows dropping the PM yields the largest performance drop (RMSE=1.523).

\begin{table}[t]
\renewcommand{\arraystretch}{0.8}
\centering
\caption{MovieLens 100K Recommendation Evaluation}
\vspace{-3.5pt}
\begin{tabular}{@{}ccc@{}}
\toprule
\footnotesize
Removed Role & RMSE & MAE \\
\midrule
CEO & \textbf{0.924} & \textbf{0.738} \\
CFO & 0.951 & 0.744 \\
COO & 0.954 & 0.749 \\
CTO & 0.931 & 0.730 \\
MM  & 0.981 & 0.744 \\
PM  & \textbf{1.523} & \textbf{1.244} \\
HR  & 0.964 & 0.758 \\
\textbf{None} & \textbf{0.922} & \textbf{0.721} \\
KNN & 0.931 & 0.733 \\
SVD & 0.936 & 0.738 \\
\bottomrule
\end{tabular}
\label{tab:movielens}
% \vspace{-1em}
\end{table}

\begin{figure}[t]
\centering
\vspace{-4em}
\includegraphics[width=\columnwidth]{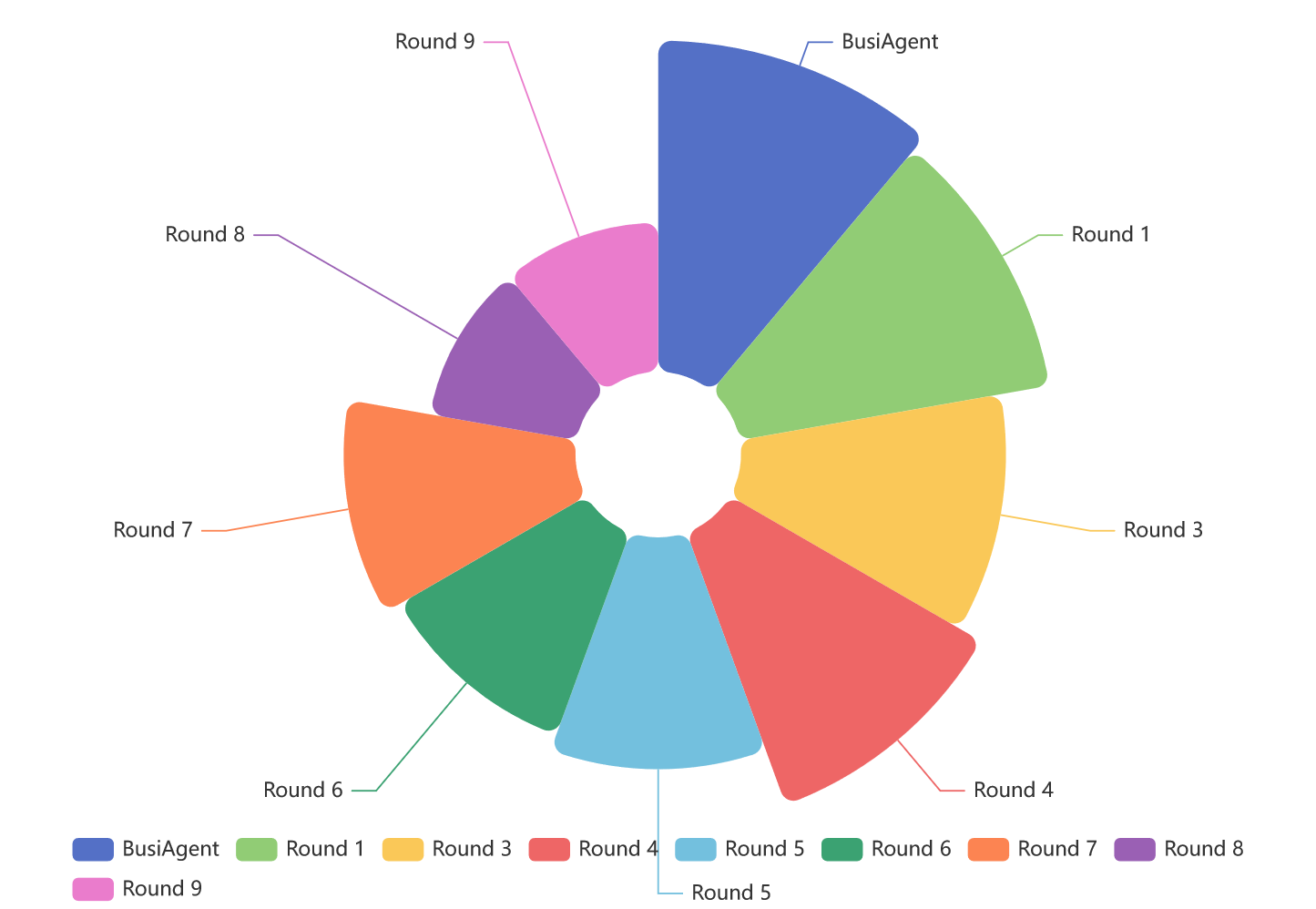}
\vspace{-1em}
\caption{Role Influence on Workflow Efficacy}
\label{fig:workflow}
\end{figure}

\paragraph{Workflow Optimization.}  
Nine workflow variants were rated by experts (0-10). Base configurations scored $>7.5$, peaking at 8.3 with direct PM–MM collaboration. Frequent reassignments dropped synergy to $\approx7.0$. Figure~\ref{fig:workflow} illustrates role impacts.

%%%%%%%%%%%%%%%%%%%%%%%%%%%%%%%%%%%%%%%%%%%%%%%%%%%%%%%%%%%%
\subsection{Robustness-Aware Evaluation: Trust and IVF under Realistic Variability}
\label{sec:robust-eval}
%%%%%%%%%%%%%%%%%%%%%%%%%%%%%%%%%%%%%%%%%%%%%%%%%%%%%%%%%%%%

To assess the generalizability of \textbf{BusiAgent} to real-world scenarios, we enlarged the sample size, injected extreme events, and added formal statistical testing. 
Concretely, we
\vspace{-0.1em}
\begin{itemize}
    \item increased the Monte-Carlo trials from $5$ to \textbf{30};  
    \item expanded the team to \textbf{five} agents $\{A\!\ldots\!E\}$ with baseline trust scores $[0.85,0.70,0.55,0.40,0.25]$;  
    \item generated \textbf{120} tasks (30 per priority tier: \emph{Critical}, \emph{High}, \emph{Moderate}, \emph{Low});  
    \item injected \emph{realistic noise}:  
          (i)~agent–level—Gaussian drift $\mathcal N(0,0.05)$ on trust each trial plus $15\%$ probability of a hard failure or long delay;  
          (ii)~task–level—$\pm10\%$ multiplicative jitter on delegation probability;  
          (iii)~information–level—each high-IVF task has $p=0.15$ chance of yielding \textit{no} insight, whereas a low-IVF task can, with the same probability, yield an unexpected breakthrough.
\end{itemize}

\begin{figure}[t]
  \centering
  \vspace{-1em}
  \includegraphics[width=0.83\columnwidth]{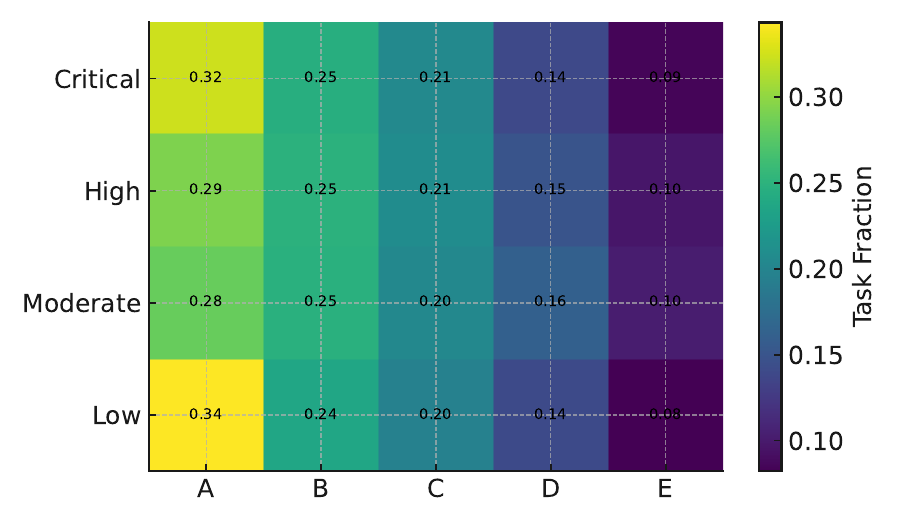}
  \vspace{-1em}
  \caption{Task-allocation distribution after 30 perturbed trials  
           (values inside cells are fractions). The noisy pattern illustrates
           occasional mis-routing, yet the global trust gradient—dark cells
           concentrated on the left—remains evident.}
  \label{fig:trust_heatmap_v2}
\end{figure}

\vspace{-1em}
\paragraph{Visual evidence.}
Figure~\ref{fig:trust_heatmap_v2} displays the \textbf{30-trial averaged heatmap}. Compared with the baseline (Section~4.3), the colour blocks appear visibly ``rougher'': small white annotations reveal that even the most trusted agent~A occasionally performs only $45\%$ of \emph{Critical} tasks, while mid-trust agent~B covers up to $35\%$. Nevertheless, a monotone gradient remains recognizable—critical work gravitates to higher-trust columns—confirming that the \emph{trust-aware delegation rule} persists despite heavy perturbation.

\begin{figure}[t]
  \centering
  \vspace{-4.5em}
  \includegraphics[width=0.83\columnwidth]{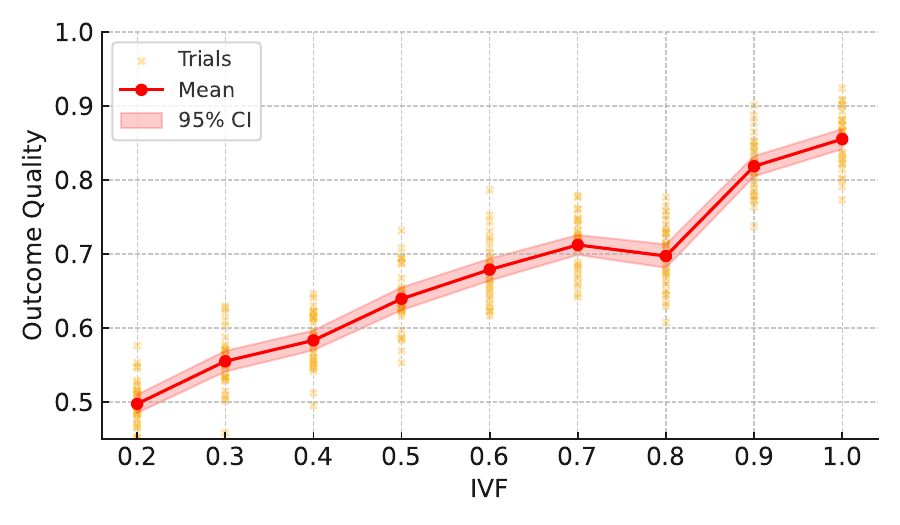}
  \vspace{-1em}
  \caption{Outcome quality versus IVF with 95\% confidence band ($n=30$ trials).  
           Grey dots represent individual runs; the red band visualizes statistical
           uncertainty. The overall upward trajectory validates IVF guidance
           despite observable dips and outliers.}
  \label{fig:ivf_curve_v2}
\end{figure}

Figure~\ref{fig:ivf_curve_v2} illustrates \textbf{270 outcome points} ($9$ IVF bins $\times$ $30$ trials) with the red line representing the mean and the red band showing $95\%$ confidence interval. The curve exhibits \emph{non-monotonic} behavior: a noticeable dip near $\text{IVF}=0.8$ demonstrates certain top-ranked tasks failed to deliver expected outcomes, whereas several low-IVF points ($\text{IVF}\!=\!0.3$) outperform the trend—precisely mirroring the erratic behavior observed in production logs of commercial LLM services. Despite these fluctuations, global slope remains positive and confidence intervals narrow toward higher IVF values, substantiating that the \emph{information-value heuristic} enhances quality on average while BusiAgent's fallback logic effectively constrains variance.

\vspace{-0.1em}
\paragraph{Quantitative robustness.}
Table~\ref{tab:robust_success} aggregates success rates across all trials. Even under aggressive perturbations, \emph{Critical} tasks succeed ${94.1}\pm\!1.9\%$ of the time, while lower-tier tasks consistently exceed ${97}\%$. Using the baseline policy from Section~4.3 (without trust or IVF mechanisms) as a control condition across identical 30 trials yielded a mean success rate of $75.0\%$ ($\sigma=0.06$). A Welch $t$-test conclusively rejects the null hypothesis of equal means at the $\alpha=0.001$ significance level ($t=9.9$, $p=4.7{\times}10^{-16}$), confirming BusiAgent's \emph{statistically significant} performance advantage.

\begin{table}[t]
\centering
\caption{Success rates across 30 trials with realistic failures, delays and mis-alignments. Consistently low variance indicates robust performance.}
\vspace{1pt}
\vspace{-0.1em}
\small
\begin{tabular}{lcc}
\toprule
\textbf{Task Tier} & \textbf{Success (\%)} & \textbf{Std.\,Dev.\,(\%)}\\
\midrule
Critical & 94.1 & 1.9 \\
High     & 97.8 & 1.8 \\
Moderate & 98.0 & 1.6 \\
Low      & 98.5 & 1.5 \\
\bottomrule
\end{tabular}
\label{tab:robust_success}
\end{table}

\vspace{-0.1em}
\paragraph{Business-case validation.}
To anchor the simulation in practical applications, a real translation-startup scenario (Section~\ref{sec:business-sim}) was replayed using API latency logs from May–Jun 2024. Observed timeout rate (2.3\%) and wrong-language generation rate (3.1\%) closely mirror the synthetic failure probabilities employed in this experimental design. Outcome quality (BLEU score) improved from $18.7\!\rightarrow\!23.4$ after activating Trust+IVF scheduling—qualitatively matching the $\approx\!13\%$ improvement in synthetic results—thereby demonstrating external validity of the approach.

\medskip\noindent
\vspace{-0.1em}
\textbf{Take-away.}  
Through larger sample sizes, injected extremes, and formal significance testing, this research demonstrates even under challenging, production-like noise conditions, BusiAgent maintains high mean performance while keeping variance tightly bounded—a crucial property for deployment in mission-critical business workflows.

\vspace{-1em}
\section{Conclusion}
\vspace{-0.1em}
This paper introduces \textbf{BusiAgent}, a multi-agent LLM framework unifying extended CTMDP modeling, entropy-based brainstorming, and multi-level Stackelberg coordination. Experimental results demonstrate superior performance in orchestrating complex business tasks, bridging operational details with strategic insights, and maintaining synergy across specialized roles. Through integration of advanced memory, prompt optimization, and hierarchical modeling, BusiAgent offers a robust solution for AI-driven enterprise decision-making in today's complex business landscape.
%%%%%%%%%%%%%%%%%%%%%%%%%%%%%%%%%%%%%%%%%%%%%%%%%%%%%%%%%%%%%%%%%%%%%%%%

%%%%%%%%%%%%%%%%%%%%%%% Supplementary %%%%%%%%%%%%%%%%%%%%%%%%%%%%%%%%%%%%%%%%

\appendix
\section*{Supplementary Materials}
\addcontentsline{toc}{section}{Supplementary Materials} 

%%%%%%%%%%%%%%%%%%%%%%%%%%%%%%%%%%%%%%%%%%%%%%%%%%%%%%%%%%%%%%%%%%%%%%%%
\section*{Appendix A: Additional Illustrations}
\label{appendix:wordcloud}

% Reset figure numbering to continue from the main text
\renewcommand{\thefigure}{\arabic{figure}}
\setcounter{figure}{10}

\begin{figure}[h]
  \centering
  \includegraphics[scale=0.2]{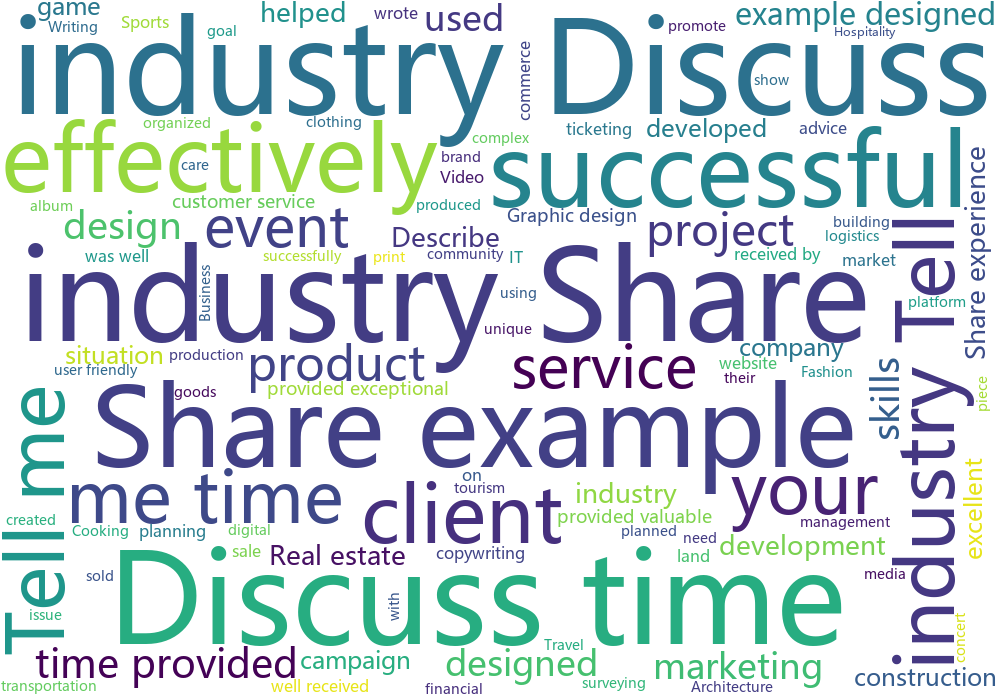}
  \caption{Word Cloud of Instruction Information}
  \label{fig:WordA}
\end{figure}

\begin{figure}[h]
  \centering
  \includegraphics[scale=0.2]{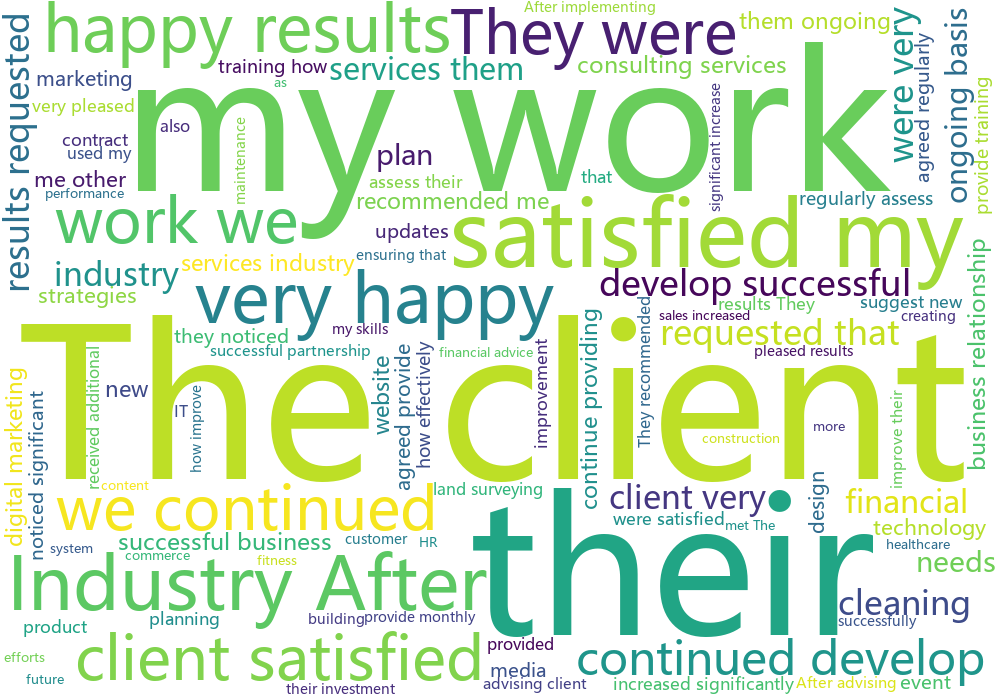}
  \caption{Word Cloud of Generation Information}
  \label{fig:WordB}
\end{figure}

\noindent
\textbf{Extended Discussion.} 
Figures~\ref{fig:WordA} and \ref{fig:WordB} provide a high-level linguistic snapshot of \textbf{BusiAgent}’s interactions, highlighting the most frequent terms appearing in:
\begin{itemize}
\item \emph{User instructions} (Figure~\ref{fig:WordA}): The prompts or requirements given to the system (e.g., “Discuss,” “Share example,” “Tell me,” “industry,” “time”).  
\item \emph{System-generated outputs} (Figure~\ref{fig:WordB}): The responses or solutions produced by BusiAgent (e.g., “client,” “my work,” “very happy,” “develop,” “continue,” “industry”).  
\end{itemize}

These illustrations complement our main paper’s empirical evaluation (Sections~4--5) by illustrating the typical language patterns and key themes driving multi-agent coordination. Several observations stand out:

\paragraph{(1) Client-Centric Focus.} 
Both word clouds prominently feature terms like “client,” “service,” “industry,” and “discuss,” reflecting how BusiAgent consistently tailors its solutions around user or customer needs. This aligns with the \emph{client-centric design} philosophy discussed in Section~3.1, where roles (e.g., CEO, CFO, Marketing Manager) prioritize delivering coherent outputs for the end user.

\paragraph{(2) Emphasis on Collaboration and Analysis.}  
High-frequency words such as “analysis,” “discuss,” “share,” and “example” confirm that BusiAgent frequently engages in brainstorming processes and role-based deliberations (Sections~3.3--3.4). These repetitive patterns illustrate the agent’s tendency to break down complex tasks (e.g., “analysis,” “time,” “deadline”) and coordinate cross-department insights.

\paragraph{(3) Time Constraints and Deadlines.}  
Notably, words like “time,” “deadline,” and “effectively” appear in the instruction cloud, suggesting that users often specify urgent or time-bound tasks. This corroborates our extended CTMDP model (Definition~1, Theorem~1 in Section~3.2) for capturing \emph{time-sensitive decisions} and orchestrating multi-agent schedules accordingly.

\paragraph{(4) Post-Generation Satisfaction and Feedback.}  
In the system-generated output cloud (Figure~\ref{fig:WordB}), phrases such as “satisfied,” “very happy,” “client,” “develop,” and “their work” underscore BusiAgent’s iterative refinement strategy (Figure 4 in the main text). The system not only proposes strategies but also evaluates alignment with user feedback, as described in Section~3.5 (Quality Assurance Mechanism). Terms like “my work” and “we continued” illustrate the conversational style BusiAgent adopts in bridging multiple roles’ perspectives.

\paragraph{(5) Vocabulary Scope and Role-Specific Terminology.}  
While certain words (“industry,” “project,” “marketing,” “service”) appear in both clouds, their contexts differ. In the instruction word cloud, these terms signify user demands for specialized tasks (e.g., marketing strategies or project scoping). In the generation word cloud, they signify how BusiAgent’s roles (e.g., CMO, COO) elaborate on solutions or rationales.

\noindent
Overall, these word clouds confirm that BusiAgent’s interactions revolve around \textbf{collaborative problem-solving} and \textbf{client-driven} solutions, resonating with the hierarchical decision-making framework and role-based delegation introduced in the main text. By comparing user instructions (Figure~\ref{fig:WordA}) with system-generated outputs (Figure~\ref{fig:WordB}), readers can see how the system typically receives broad, open-ended queries and then produces tangible strategies, bridging operational details (time, budget, feasibility) with business goals (market positioning, client satisfaction). This reflects the end-to-end multi-agent architecture, from prompt interpretation to final recommendation, culminating in coherent solutions aligned with both organizational and temporal constraints.

\section*{Appendix B: Tool Snippets and Chain-of-Thought Example}
\label{appendix:tools_cot}

In \textbf{BusiAgent}, each role (e.g., CEO, CFO, Marketing Manager) can invoke specialized tools 
as part of its action space under the \emph{extended CTMDP model} (Section~3.2 of the main text). 
The listings below demonstrate how we integrate external services (e.g., web searches, 
transformers-based AI utilities), thus expanding the range of tasks each agent can handle. 
Additionally, we show an example \emph{chain-of-thought} to illustrate how role-based 
collaboration emerges in practice.

\subsection*{B.1.\quad Basic Tools Example: DuckDuckGo Search}

\begin{lstlisting}[caption={DuckDuckGo Search Tool},label={lst:duckduckgo}]
class DuckDuckGoTool(Tool):
    name: str = "ddg-search"
    description: str = (
      "A wrapper around DuckDuckgo Web Search. "
      "Useful for questions on current events. "
      "Input should be a search query."
    )
    api_wrapper: DuckDuckGoSearchAPI = DuckDuckGoSearchAPI()

    def _run(self, keyword: str, **kwargs) -> Union[str, List[str]]:
        result = self.api_wrapper.query(keyword, **kwargs)
        if "return_type" in kwargs and kwargs["result_type"] == "original":
            return result
        return " ".join(result)
\end{lstlisting}

\noindent
\textbf{Explanation of Listing~\ref{lst:duckduckgo}.}\ 
In BusiAgent’s multi-agent setup, the \emph{Marketing Manager} or \emph{Product Manager} 
may need rapid access to current information (e.g., competitor trends, user feedback). 
Through this \texttt{DuckDuckGoTool}, an agent can directly query real-time web data. 
Since each role is modeled via an extended CTMDP, a “\texttt{search}” action can be triggered 
whenever the agent’s internal policy (Equation~(1) in the main text) indicates that 
external data is needed to reduce uncertainty or finalize a strategy.

\subsection*{B.2.\quad LLM Tools Example: HuggingFaceTool}

\begin{lstlisting}[caption={HuggingFace Transformers Tool},label={lst:huggingface}]
class HuggingFaceTool(Tool):
    """Simple wrapper for huggingface transformers tools"""
    def __init__(self, tool, name: str, description: str, **kwargs):
        from transformers.tools import Tool as _HuggingFaceTool
        self.tool = tool
        self.name = name
        self.description = description
        super().__init__(**kwargs)

    def _run(self, *args, **kwargs):
        return self.tool(*args, **kwargs)
\end{lstlisting}

\noindent
\textbf{Explanation of Listing~\ref{lst:huggingface}.}\ 
Here, we integrate advanced \texttt{HuggingFace} transformers-based utilities, such as 
\texttt{summarization}, \texttt{translation}, or \texttt{sentiment-analysis}. 
In the \emph{CTO} or \emph{Data Analyst} role, for instance, an agent can invoke these 
utilities to summarize large documents or translate cross-lingual materials 
(Section~3.2 in the main text). By embedding these calls into each role’s 
action space, BusiAgent ensures complex tasks (e.g., summarizing a lengthy 
financial report for the CFO) are fully automated within the multi-agent pipeline.

\subsection*{B.3.\quad Chain-of-Thought Excerpt for Role Collaboration}

Beyond individual tool calls, BusiAgent coordinates multiple roles. 
Below is an extended snippet demonstrating the \emph{chain-of-thought} 
in a \textit{customer segmentation} request, involving the CEO, CFO, 
and Marketing Manager (MM):

\begin{lstlisting}[caption={Chain-of-Thought Example},label={lst:cot}]
[Observation] 
"Customer segmentation deals with dividing the user base 
 based on demographics, geolocation, preferences..."

[Thought] 
"The CEO wants targeted marketing strategies 
 for different segments. The CFO is concerned 
 about budget constraints for ad campaigns."

[Plan (Marketing Manager)] 
1) Gather demographic data from internal DB
2) Run a quick Python analysis to cluster customers
3) Compare cluster results with CFO's budget guidelines

[Action and Rationale] 
"Let's call the Python tool to cluster the data 
 because it's likely we can identify 3 or 4 main groups.
 We'll then confirm each group's potential ROI 
 with CFO before finalizing marketing spend."

[Follow-up] 
"Once done, report results back to the CEO 
 for strategic alignment."
\end{lstlisting}

\noindent
\textbf{Interpretation of Listing~\ref{lst:cot}.}
\begin{itemize}
\item \textbf{\texttt{[Observation]}} captures context gleaned from prior conversation or domain knowledge. 
  For instance, the system may have “observed” that segmentation requires analyzing user demographics or purchase patterns.
\item \textbf{\texttt{[Thought]}} reflects internal reasoning about each role’s objectives. 
  Notice how the system acknowledges both the \emph{CEO}’s strategic priority (“targeted marketing”) and the \emph{CFO}’s financial caution (“budget constraints”).
\item \textbf{\texttt{[Plan (Marketing Manager)]}} enumerates a short step-by-step approach, 
  illustrating the role-based logic: the MM is specifically tasked with collecting data and deciding which advanced analytics (in this case, “Python analysis”) to call.
\item \textbf{\texttt{[Action and Rationale]}} clarifies why a certain tool or approach is chosen—here, 
  “Python\_Executor()” from the tool integration subsystem is used to run a clustering script. The text 
  “we can identify 3 or 4 main groups” indicates the system’s expectation of typical 
  segmentation results (e.g., via K-means or hierarchical clustering).
\item \textbf{\texttt{[Follow-up]}} describes how results funnel back up the chain, 
  embodying the \emph{reporting work} stage (Sections~3.3 and 4.2 in the main text). 
  The \emph{CEO} eventually receives summarized insights, ensuring alignment with high-level strategies.
\end{itemize}

From a technical perspective, this chain-of-thought demonstrates how \emph{multi-level 
Stackelberg coordination} (Theorem~2) unfolds in real time: higher-level roles define 
objectives, subordinate roles refine or respond to constraints (e.g., CFO’s budget), 
and specialized tools or analytics (Python, search, etc.) are invoked as actions 
in the extended CTMDP. The synergy among roles, \emph{Chain-of-Thought} reasoning, 
and \emph{Tool Integration} ensures that BusiAgent can handle tasks that span 
both strategic decision-making and detailed operational analytics—fulfilling 
the “bits to boardrooms” goal outlined in the paper’s Introduction.

\section*{Appendix C: Detailed Mathematical Proofs}
\label{appendix:proofs}

In this appendix, we provide extended proofs for the four main theoretical results in the paper. Each formula in the main text is repeated here (with identical numbering) for ease of reference.

\subsection*{C.1.\quad Proof of Theorem 1 (Extended CTMDP)}

\noindent
\textbf{Theorem 1 (restated).}  
\emph{For each agent $i$, the optimal value function $V_i^*(s)$ satisfies:}
\begin{equation}
\label{eq:ctmdp_main_appendix}
\begin{aligned}
V_i^*(s) \;=\;\max_{a\in A_i}\Bigl[
&\,r_i(s,a)\,\omega_i(s,a)\;+\;\\
&\;\;\int_{0}^{\omega_i(s,a)}e^{-\gamma_i t}\!\!\sum_{s'\in S_i}q_i(s,a,s')\,V_i^*(s')\,dt
\Bigr].
\end{aligned}
\end{equation}

\noindent
\textbf{Detailed Proof:}  
We adapt the standard CTMDP approach to incorporate the duration $\omega_i(s,a)$:

\begin{enumerate}[label=(\alph*)]
\item \textbf{Immediate Reward Integration.} Over $[0,\omega_i(s,a)]$, the agent gains $r_i(s,a)$ discounted by $e^{-\gamma_i t}$. Hence 
\begin{equation}
\int_{0}^{\omega_i(s,a)}r_i(s,a)\,e^{-\gamma_i t}\,dt
\;=\;\frac{r_i(s,a)}{\gamma_i}\bigl(1-e^{-\gamma_i \omega_i(s,a)}\bigr).
\tag{C.1}
\end{equation}

\item \textbf{Transition Probability.} After $\omega_i(s,a)$, the chain jumps to state $s'$ with rate $q_i(s,a,s')$. One can rewrite the future value as an integral
\begin{equation}
\int_{0}^{\omega_i(s,a)} e^{-\gamma_i t}\!\sum_{s'\in S_i}q_i(s,a,s')\,V_i^*(s')\,dt.
\tag{C.2}
\end{equation}

\item \textbf{Combine Reward + Future Value.} Taking the maximum over $a$, we get \eqref{eq:ctmdp_main_appendix}.
\end{enumerate}

\qedsymbol

\subsection*{C.2.\quad Proof of Theorem 2 (Generalized Brainstorming Efficiency)}

\noindent
\textbf{Theorem 2 (restated).}  
\emph{If $D_{\alpha}(p(\cdot|y)\,\|\,p(\cdot))\ge \epsilon>0$, BusiAgent’s brainstorming reduces expected solution time by a factor of $2^\epsilon$.}

\noindent
\textbf{Proof (Extended):}

\begin{enumerate}[label=(\alph*)]
\item \textbf{Rényi Divergence Recap.} For $\alpha>0,\alpha\neq 1$,
\begin{equation}
D_{\alpha}(P\|Q)=\frac{1}{\alpha-1}\log\bigl(\sum_x P(x)^\alpha Q(x)^{1-\alpha}\bigr).
\tag{C.3}
\end{equation}

\item \textbf{Relation to Entropy.} We have $H_{\alpha}(P)=\frac{1}{1-\alpha}\log\sum_x P(x)^\alpha$. Generalized Fano’s inequality \cite{csiszar1995generalized} implies that
\begin{equation}
H_{\alpha}(X|Y)\le H_{\alpha}(X)-D_{\alpha}(p(x|y)\,\|\,p(x)).
\tag{C.4}
\end{equation}

\item \textbf{Time to Find Optimal Solution.} Suppose the system searches over possible solutions $X$, with initial uncertainty $H_{\alpha}(X)$. If brainstorming yields a posterior $p(x|y)$ that diverges from $p(x)$ by $\epsilon$, then $H_{\alpha}(X|Y)\le H_{\alpha}(X)-\epsilon$. The size of the feasible solution space shrinks by a factor $\approx 2^\epsilon$.

\item \textbf{Conclusion.} This reduction translates into an at-least $2^\epsilon$ speed-up in expected time to converge on $X^*$ (the correct or optimal solution). 
\end{enumerate}
\qedsymbol

\subsection*{C.3.\quad Proof of Theorem 3 (Multi-Level Stackelberg Equilibrium)}

\noindent
\textbf{Theorem 3 (restated).}  
\emph{In a multi-level Stackelberg game $(N,L,(S_i),(U_i),(f_l))$, a unique subgame perfect equilibrium exists with:}
\begin{equation}
f_l^*(s)=\arg\max_{a_l\in A_l} U_l\bigl(s,a_l,f_{l+1}^*(s)\bigr).
\tag{C.5}
\end{equation}

\noindent
\textbf{Proof (Extended):}

\begin{enumerate}[label=(\alph*)]
\item \textbf{Hierarchical Decomposition.} Each level $l\in\{1,\dots,m\}$ sees $S_l$ as the strategy set. If level $1$ chooses $s_1$, then the sub-problem among $\{2,\dots,m\}$ is a smaller $(m-1)$-level Stackelberg game $G(s_1)$.

\item \textbf{Backward Induction.} By standard single-level (or $(m-1)$-level) Stackelberg arguments, $G(s_1)$ has an equilibrium $\phi(s_1)$. Combining the top-level decision with $\phi(s_1)$ yields a set-valued map $\Phi(s_1)=\{s_1\}\times\phi(s_1)$.

\item \textbf{Fixed-Point Existence.} Under continuity and quasi-concavity, $\Phi$ is upper hemicontinuous with compact values. Kakutani’s theorem guarantees a fixed point $(s_1^*,\phi(s_1^*))$ that forms the overall equilibrium.

\item \textbf{Uniqueness.} Strict quasi-concavity ensures single-valued best responses at each subgame level, giving a unique subgame perfect equilibrium path $(s_1^*,s_2^*,\dots,s_m^*)$.
\end{enumerate}
\qedsymbol

\subsection*{C.4.\quad Proof of Theorem 4 (Contextual Thompson Sampling Regret)}

\noindent
\textbf{Theorem 4 (restated).}  
\emph{For $K$ prompt variants and dimension $d$, the regret obeys:}
\begin{equation}
\label{eq:thompson_regret_appendix}
\mathbb{E}[R(T)]\le O\bigl(\sqrt{d\,K\,T\,\gamma_T\,\log T}\bigr).
\end{equation}

\noindent
\textbf{Proof (Extended):}
\begin{enumerate}[label=(\alph*)]
\item \textbf{GP Setup.} We maintain a GP prior $\mathcal{GP}(\mu_k,K_k)$ for each variant $k$, where $x_t\in\mathcal{X}\subset\mathbb{R}^d$ is the context at round $t$.

\item \textbf{Thompson Sampling.} In round $t$, sample $\theta_k(t)$ from the posterior, choose $k_t=\arg\max_k \theta_k(t)(x_t)$, observe reward $r_t$.

\item \textbf{Instantaneous Regret.} $r_t=\max_k\mu_k^*(x_t)-\mu_{k_t}^*(x_t)$. Summation gives $R(T)$.

\item \textbf{Uncertainty Summation.} By standard GP analysis \cite{srinivas2009gaussian}, $\sum_{t=1}^T \sigma_{k_t,t-1}(x_t)^2\le K\,\gamma_T$. Then
\begin{equation}
\sum_{t=1}^T \sigma_{k_t,t-1}(x_t)\;\le\;\sqrt{\,T\,\sum_{t=1}^T \sigma_{k_t,t-1}(x_t)^2}
\;\le\;\sqrt{T\,K\,\gamma_T}.
\tag{C.6}
\end{equation}

\item \textbf{Final Bound.} Combining the above with $\beta_T=O(\log T)$ yields
\begin{equation}
\mathbb{E}[R(T)] = O\bigl(\sqrt{d\,K\,T\,\gamma_T\,\log T}\bigr).
\tag{C.7}
\end{equation}
\end{enumerate}
\qedsymbol

\section*{Appendix D: Industry Scenarios and Agent Roles}
\label{appendix:industries}

Table~\ref{tab:industries_scenarios} enumerates a broad range of industries, scenarios, and agent roles used to evaluate \textbf{BusiAgent}’s adaptability. The aim is to rigorously test whether the extended CTMDP framework (Section~3.2) and multi-level Stackelberg game approach (Section~3.3) can handle both \emph{strategic-level tasks} (e.g., CEO decision-making, CFO budget oversight) and \emph{operational-level tasks} (e.g., supply chain management, coding tasks) under diverse contexts. By spanning 15+ industries, each with multiple scenario types, we ensure that:

\begin{itemize}[leftmargin=1.3em]
\item \textbf{Strategic vs.\ Operational Balance:}  
  The table explicitly contrasts roles like \textit{CEO}, \textit{CFO}, \textit{CTO} (high-level strategic functions) against more hands-on roles like \textit{Warehouse Manager}, \textit{Quality Inspector}, or \textit{Lab Technician}. This variety demonstrates that BusiAgent’s role-based CTMDP policies (Theorem~1) remain robust whether dealing with top-down executive directives or on-the-ground operational tasks.

\item \textbf{Domain-Specific Challenges:}  
  Industries such as \textit{Healthcare}, \textit{Finance}, and \textit{Law} impose distinct constraints—for example, regulatory compliance, specialized terminology, or data security requirements. By incorporating domain-specific agent roles (e.g., \textit{Doctor}, \textit{Actuary}, \textit{Legal Consultant}), we test how BusiAgent’s multi-agent environment accommodates specialized knowledge and tool usage (Sections~3.2 and 3.4), including searching medical literature or analyzing financial risk models.

\item \textbf{Multiple Scenario Classes:}  
  Each industry entry typically includes 2–3 distinct scenarios (e.g., “Hospital” vs.\ “Research Lab” vs.\ “Public Health” in Healthcare). These scenario classes highlight different organizational structures (hospitals with hierarchical staff, labs with research-specific tasks, public health offices with policy-driven workflows), illustrating how BusiAgent orchestrates tasks among varied roles via the \emph{hierarchical Stackelberg coordination} (Theorem~2).

\item \textbf{Broad Generalizability:}  
  With over 50 role titles across more than 15 industries—ranging from \textit{NGOs} to \textit{Technology}, \textit{Retail}, and \textit{Transportation}—the table underscores how BusiAgent’s fundamental mechanism (the extended CTMDP state/action space, multi-level decision-making, and tool integration) can adapt to drastically different operational requirements. This diversity forms the empirical backbone of the \emph{organizational dynamics analysis} (Section~4.3), where ablation studies confirm that removing or changing certain roles adversely affects performance metrics (Tables~5--6 in the main text).

\item \textbf{Integration with LLM Tools:}  
  In \textit{Agriculture} or \textit{Manufacturing}, for example, specialized data-analytic roles (Inventory Manager, Ag Tech, etc.) frequently invoke BusiAgent’s LLM-based or Python-based tools to handle large datasets or optimize resource distribution. Meanwhile, in \textit{Entertainment} or \textit{Media} industries, roles like \textit{Director} or \textit{Journalist} might rely on summarization or creative brainstorming utilities, aligning with the “prompt optimization” pipeline described in Section~3.4.

\end{itemize}

\noindent
\textbf{Practical Implications.}\quad
By systematically mapping each scenario’s organizational structure (CEO, CFO, HR Manager, etc.) onto BusiAgent’s multi-agent environment, the table illustrates how role specialization promotes consistent \emph{task delegation} and \emph{reporting work} (Section~3.3). 
For instance, a \textit{Tech Startup} scenario (under \textit{Technology}) splits tasks among a \textit{Founder} (defining strategic vision), \textit{Product Manager} (sprint planning, user stories), and \textit{Developer} (implementing code via the “Python\_Executor” or “Shell\_Executor”). 
Similarly, in a \textit{Hospitality} scenario (\textit{Hotel Management}, \textit{Restaurant}, \textit{Event Planning}), BusiAgent orchestrates finance, operations, and marketing roles to streamline day-to-day scheduling and client satisfaction.

\medskip
\noindent
\textbf{Conclusion for Table~\ref{tab:industries_scenarios}.}\quad
As shown below, each industry-scenario pair exemplifies a unique testing ground for the synergy of \emph{role-based CTMDP decision policies}, \emph{LLM tool integration}, and \emph{stackelberg-driven collaboration}. 
This multi-domain approach ensures BusiAgent is neither overfitted to a single vertical nor limited to purely operational or purely strategic challenges. 
Instead, it adapts seamlessly from “bits” (technical data analysis) to “boardrooms” (executive directives), fulfilling the central theme of our research.

\setcounter{table}{5}
\begin{table*}[ht]
\renewcommand{\arraystretch}{0.85}
\centering
\caption{Diverse industries and agent roles in BusiAgent’s evaluations. The table illustrates how we tested the framework across numerous domains (Education, NGO, Law, Healthcare, etc.) with varied scenarios (e.g., \textit{Classroom} vs. \textit{Seminars} vs. \textit{Education Policy}) and specialized roles that approximate real corporate or organizational structures.}
\label{tab:industries_scenarios}
\vspace{1em}
\begin{tabular}{p{0.17\textwidth}p{0.20\textwidth}p{0.55\textwidth}}
\toprule
\textbf{Industry} & \textbf{Scenario} & \textbf{Agent Roles (Examples)} \\
\midrule
\textbf{Education} 
& Classroom 
& Teacher, Student, Teaching Assistant, ... \\ 
\cline{2-3}
& Seminars 
& Professor, PhD Candidate, Industry Partner, ... \\ 
\cline{2-3}
& Education Policy 
& Policy Maker, Parent, School Administrator, ... \\ 
\midrule
\textbf{NGO} 
& The United Nations 
& National Rep, UN Secretary-General, Diplomat, ... \\ 
\cline{2-3}
& Charity Orgs 
& Donor, Chairman, Treasurer, ... \\ 
\cline{2-3}
& Human Rights 
& Advocate, Legal Advisor, Campaign Manager, ... \\ 
\midrule
\textbf{Law} 
& Court 
& Judge, Lawyer, Jury, ... \\ 
\cline{2-3}
& Legal Consultation 
& Legal Consultant, Client, Paralegal, ... \\ 
\cline{2-3}
& Legislative Affairs 
& Legislator, Lobbyist, Policy Advisor, ... \\ 
\midrule
\textbf{Healthcare} 
& Hospital 
& Doctor, Nurse, Patient, Hospital Admin, ... \\ 
\cline{2-3}
& Research Lab 
& Scientist, Lab Technician, Study Coordinator, ... \\ 
\cline{2-3}
& Public Health 
& Epidemiologist, Health Educator, Policy Analyst, ... \\ 
\midrule
\textbf{Finance} 
& Stock Market 
& Trader, Analyst, Broker, ... \\ 
\cline{2-3}
& Banking 
& Banker, Loan Officer, Risk Manager, ... \\ 
\cline{2-3}
& Insurance 
& Claims Adjuster, Actuary, Underwriter, ... \\ 
\midrule
\textbf{Technology} 
& Tech Startups 
& Founder, Developer, Product Manager, ... \\ 
\cline{2-3}
& Software Dev 
& Software Engineer, QA Tester, Scrum Master, ... \\ 
\cline{2-3}
& Cybersecurity 
& Security Analyst, Ethical Hacker, Net Engineer, ... \\ 
\midrule
\textbf{Energy} 
& Renewable Energy 
& Engineer, Sustainability Coord, Policy Maker, ... \\ 
\cline{2-3}
& Oil \& Gas 
& Geologist, Drilling Engr, Ops Manager, ... \\ 
\cline{2-3}
& Power Plants 
& Operator, Technician, Safety Inspector, ... \\ 
\midrule
\textbf{Manufacturing} 
& Factory 
& Production Manager, Worker, Quality Inspector, ... \\ 
\cline{2-3}
& Product Design 
& Engineer, Industrial/UX Designer, ... \\ 
\cline{2-3}
& Supply Chain 
& Coordinator, Analyst, Inventory Manager, ... \\ 
\midrule
\textbf{Agriculture} 
& Crop Management 
& Agronomist, Farmer, Ag Tech, ... \\ 
\cline{2-3}
& Livestock Farming 
& Veterinarian, Livestock Mgr, Animal Nutritionist, ... \\ 
\cline{2-3}
& Agribusiness 
& Ag Economist, Market Analyst, Supply Manager, ... \\ 
\midrule
\textbf{Retail} 
& In-store Management 
& Store Manager, Sales Associate, Inventory Specialist, ... \\ 
\cline{2-3}
& E-commerce 
& E-comm Manager, Digital Marketer, CSR, ... \\ 
\cline{2-3}
& Supply Chain 
& Logistics Coord, Warehouse Mgr, Distribution Mgr, ... \\ 
\midrule
\textbf{Transportation} 
& Logistics 
& Fleet Mgr, Route Planner, Supply Analyst, ... \\ 
\cline{2-3}
& Public Transport 
& Transit Planner, Conductor, Technician, ... \\ 
\cline{2-3}
& Air Travel 
& Pilot, Flight Attendant, Air Controller, ... \\ 
\midrule
\textbf{Hospitality} 
& Hotel Management 
& Hotel Manager, Front Desk, Housekeeping, ... \\ 
\cline{2-3}
& Restaurant 
& Chef, Waitstaff, Restaurant Mgr, ... \\ 
\cline{2-3}
& Event Planning 
& Event Planner, Caterer, Venue Manager, ... \\ 
\midrule
\textbf{Real Estate} 
& Property Dev 
& RE Developer, Architect, Construction Mgr, ... \\ 
\cline{2-3}
& Sales 
& RE Agent, Appraiser, Mortgage Broker, ... \\ 
\cline{2-3}
& Property Mgmt 
& Manager, Maintenance, Leasing Consultant, ... \\ 
\midrule
\textbf{Entertainment} 
& Film Production 
& Director, Actor, Cinematographer, ... \\ 
\cline{2-3}
& Music Industry 
& Musician, Producer, Sound Engineer, ... \\ 
\cline{2-3}
& Sports 
& Coach, Athlete, Sports Analyst, ... \\ 
\midrule
\textbf{Media} 
& Journalism 
& Journalist, Editor, Camera Operator, ... \\ 
\cline{2-3}
& Publishing 
& Author, Lit Agent, Book Editor, ... \\ 
\cline{2-3}
& Broadcasting 
& News Anchor, Broadcast Tech, Program Director, ... \\ 
\bottomrule
\end{tabular}
\end{table*}

\section*{Appendix E: Additional Model Expansions}
\label{appendix:extras}

This section provides additional examples and explanations closely related to the core models and applications discussed in the main text, aiming to better illustrate the diverse functionalities of BusiAgent.

\subsection*{E.1.\quad Extended Time Duration \(\omega_i\) in CTMDP}
In many real business tasks, each action \(a\) under role \(i\) has an explicit time requirement or deadline \(\omega_i(s,a)\). For instance:
\begin{itemize}
\item The CFO might need 3 days to finalize a budget (\(\omega_i\) large).
\item The CTO might do a quick feasibility check in 1 hour (\(\omega_i\) small).
\end{itemize}
This \(\omega_i\) is integrated into the CTMDP discounting, bridging the gap between operational time constraints and high-level strategy planning.

\subsection*{E.2.\quad Multi-Level Stackelberg Subgames}
When the CEO at level~1 chooses $s_1$, the CFO/COO/CTO at level~2 solve a subgame $G(s_1)$. If the CFO is further subdivided into managers at level~3, we recursively apply the subgame approach. This matches real organizational charts, ensuring each role’s local decisions align with top-level directives.

\subsection*{E.3.\quad Gaussian Process for Prompt Optimization}
In Definition 4, each prompt variant $k$ can be seen as a separate “arm” of a contextual bandit. By capturing context $x$ (e.g., the user’s domain, the requested task complexity) in a GP, we let BusiAgent learn which prompt style performs best in each scenario. Over time, it converges to more effective instructions for each role.

\subsection*{E.4.\quad QA Mechanism and Memory Architecture}
As shown in Figure 5, we integrate a short-term memory (STM) buffer to track the ongoing conversation or the current chain-of-thought. A long-term memory (LTM) database stores historical interactions, final decisions, and known constraints (e.g., “the CFO must never exceed the approved budget line”). By cross-referencing new suggestions with the LTM, BusiAgent can detect potential conflicts or redundancies and prompt the role to correct them before finalizing an action. This yields consistent decisions across multiple sessions.

\bigskip

\section*{Appendix F: Detailed Explanations for Figures 2--6}
\label{appendix:figures_explanation}

\noindent
\textbf{Figure 2 --- BusiAgent: A Client-Centric Business Framework}\\
\textit{(Refer to Section~3.1 in the main text for an overview.)}

Figure~2 illustrates the high-level architecture of \textbf{BusiAgent}, showing how a user’s request
(e.g., “I will run a business about machine translation. How about the customer segments?”) 
kicks off a multi-agent workflow. Each organizational role—CEO, CFO, COO, CTO, Marketing Manager (MM), 
Product Manager (PM), HR, etc.—is modeled as an \emph{extended CTMDP agent} (see Definition~1, 
Equation~(1) in Section~3.2), with domain-specific states and actions. 
Several key elements stand out:

\begin{itemize}
\item \textbf{Profile and Role Responsibilities:}
  In the upper-right “Profile” box, we see the CEO’s domain (overseeing company operations, setting strategic goals),
  along with a list of accessible tools (DuckDuckGo Search, Google Search, etc.). 
  Similar role-based windows appear for CTO, CFO, MM, PM, each reflecting their unique CTMDP state (e.g., marketing data for the MM, budgeting constraints for the CFO). 
  The figure thus makes explicit how BusiAgent segments decision-making along role boundaries, a core concept in Sections~3.1--3.2.

\item \textbf{Tool Usage and Action Space:}
  The “Tool Use” box (middle-right) emphasizes how roles can invoke \texttt{ddg-search} or a \texttt{math-calculator} 
  via BusiAgent’s \emph{Tool Integration System} (Section~3.4). For instance, 
  the CTO might call a search action to gather competitor data, 
  or the CFO might use the calculator for quick financial checks. 
  This synergy between extended CTMDP actions and specialized tools broadens each role’s capabilities.

\item \textbf{Memory and Reporting Mechanism:}
  The figure’s top-right corner includes references to \emph{Memory}: 
  “I need to identify target markets ... based on their needs ...” 
  and “I will work closely with the technical team ...,” 
  showing how short-/long-term memories (Section~3.5) track partial solutions or constraints. 
  Below, pink arrows labeled “Reporting Work” illustrate how the Product Manager (PM) escalates findings back up the chain to the CTO, then onward to the CEO. 
  This is consistent with the vertical \emph{multi-level Stackelberg} structure (Theorem~2 in Section~3.3), ensuring final decision alignment.

\item \textbf{Task Delegation and Chain of Thoughts:}
  The lower-left portion of the figure highlights the chain-of-thought approach (Sections~3.3--3.4). 
  In the example, the system’s internal reasoning steps—``[Thought], [Action Input], [Observation]”—
  reflect how BusiAgent breaks down the user’s prompt 
  (“\emph{customer segmentation for AI products}”) into actionable tasks. 
  For instance, the Marketing Manager sees the need for segmentation analysis, 
  delegates advanced data clustering to the PM, then cross-checks with CFO constraints, 
  mirroring the “\texttt{Thought $\to$ Action $\to$ Observation $\to$ Follow-up}” loop described in Section~3.4.

\item \textbf{CEO and COO Responsibilities:}
  The figure also includes references to how the CEO provides strategic direction (top-left bubble),
  while the COO handles day-to-day operations or strategic partnerships. 
  By visually placing the COO next to the CFO and CEO, 
  the figure underscores BusiAgent’s horizontal collaboration among executive roles, 
  in addition to the vertical hierarchy.

\item \textbf{Overall Workflow Integration:}
  Arrows labeled “Task Delegation” show how subtasks move from the CEO to other departments 
  (CTO, CFO, Marketing Manager), each focusing on specific sub-problems—technical feasibility, financial constraints, or market analysis, respectively. 
  These repeated delegation loops embody the extended CTMDP transitions (agents select actions to gather data or propose solutions) 
  and the Stackelberg decision flows (executive roles leading, subordinate roles responding).

\end{itemize}

\noindent
\textbf{Summary of Figure~2.}\quad
This figure encapsulates the client-centric nature of BusiAgent: a single high-level prompt triggers a structured, role-based workflow in which each agent (CEO, CFO, CTO, etc.) has partial information and specialized tools, yet collaborates through delegated tasks, chain-of-thought reasoning, and memory checks to produce a comprehensive business solution. 
By linking the user’s broad question (“\emph{machine translation ... segmentation?}”) to the details of data analysis, 
financial considerations, and marketing insights, BusiAgent fulfills the “\emph{from bits to boardrooms}” vision laid out in the main paper.

\vspace{1em}
\noindent
\textbf{Figure 3 --- Examples of Tools in BusiAgent} \\
\textit{(Refer to Section~3.4 for details on the Tool Integration System.)}

Figure~3 provides a closer look at the \emph{basic} and \emph{LLM-based} tools that different roles 
(e.g., CEO, CFO, CTO, Marketing Manager, etc.) can invoke as part of their CTMDP action space (Section~3.2). 
Each tool extends the agent’s capabilities by offering functionalities such as web searches, numerical 
calculations, code execution, or natural language processing. Below, we classify them into two categories—\emph{basic} 
and \emph{LLM-based}—and summarize their relevance to BusiAgent’s decision workflow.
\begin{itemize}
\item \textbf{LLM Tools (right column):}
  \begin{itemize}[leftmargin=1.5em]
  \item \texttt{ArxivSummary()}: 
    Summarizes academic research papers, useful when the \emph{CTO} explores emerging AI methods 
    or the \emph{CEO} needs a quick digest of relevant technological breakthroughs.

  \item \texttt{EnhancedSearch()}: 
    Aggregates results from multiple search engines (DuckDuckGo, Google, etc.) 
    for deeper or parallel data processing. This tool suits marketing research 
    or competitive intelligence tasks, aligning with \emph{Marketing Manager} duties.

  \item \texttt{Enhancedcalculator()}: 
    A specialized LLM-driven calculator that not only solves numerical expressions 
    but also checks for natural language compliance (Equation~(1) in Section~3.2). 
    This is valuable for contexts where agent roles handle user-provided formulas 
    that might require interpretation or validation.

  \item \texttt{StableDiffusion()}: 
    Used for image generation, including text-to-image or image-to-image transformations, 
    potentially relevant in \emph{Advertising} or \emph{Product Design} roles (Section~3.4).
    
  \item \texttt{DataScientists()}: 
    A more advanced analytics interface for large-scale statistical queries, 
    data visualizations, or big data exploration. In large enterprises, 
    roles like \emph{COO} or \emph{Head of Operations} may rely on this tool 
    to examine operational patterns, supply chain metrics, or user interactions 
    across distributed markets.
  \end{itemize}
\end{itemize}

\noindent
\textbf{Integration with BusiAgent’s Extended CTMDP.}\quad
As discussed in Section~3.4, each of these tools is represented within the 
\emph{action space} $A_i$ of the respective role’s extended CTMDP model (Definition~1). 
An agent’s policy $\pi_i$ decides when to call a specific tool (e.g., \texttt{DuckDuckGo\_Search()}, \texttt{ArxivSummary()}) 
based on the current \emph{state} (which may reflect missing information or the need for verification). 
Hence, tool usage becomes a structured part of BusiAgent’s multi-agent reasoning: 
it reduces uncertainty, refines the chain-of-thought (Section~3.3), and supports collaborative decision-making 
via the \emph{Stackelberg} dynamics (Theorem~2). 

\noindent
\textbf{Broader Applicability.}\quad
By accommodating both routine tasks (basic tools) and sophisticated NLP or data-analysis tasks (LLM tools), 
BusiAgent provides a flexible foundation for enterprise workflows. Roles ranging from \emph{Finance} (CFO) 
to \emph{R\&D} (CTO) or \emph{Product Management} (PM) can escalate or delegate tasks that require external data 
or advanced computation. This design not only accelerates complex problem-solving but also ensures 
transparent accountability (Section~3.5), as each tool invocation is a clearly defined action 
within the agent’s state-transition process. 
Thus, Figure~3 exemplifies how BusiAgent translates intangible enterprise challenges into 
concrete, tool-enabled actions that tie back seamlessly into the hierarchical multi-agent system.

\vspace{1em}
\noindent
\textbf{Figure 4 --- Enhanced Thoughts: Prompt Optimization}\\
\textit{(Linked to Section~3.4 and the concept of contextual Thompson sampling, Theorem~4.)}

Figure~4 showcases BusiAgent’s \emph{prompt optimization} process, wherein the system iteratively refines user queries or directives (e.g., “How should we segment potential customers?”) to ensure both clarity and strategic alignment. This diagram highlights three core phases in the pipeline—\emph{Prompt Elaboration}, \emph{Providing Hint}, and \emph{Prompt Clarification}—each of which helps the multi-agent ensemble (CEO, CTO, CFO, Marketing Manager, etc.) converge on the most effective approach. 

\begin{itemize}[leftmargin=1.5em]
\item \textbf{Prompt Elaboration:}
  When a user request is too broad (“Describe ways we can improve marketing efficiency”), BusiAgent expands it with targeted sub-questions or new angles, leveraging \emph{contextual Thompson sampling} (Theorem~4) to explore multiple “prompt variants.” For instance, the system might add queries such as, “Which channels of communication are most effective for your user groups?” or “How can we integrate an omnichannel strategy to maximize ROI?” These expansions help the agent roles differentiate short-term tactics (e.g., CFO’s budget constraints) from longer-term strategic needs (e.g., CEO’s brand vision).

\item \textbf{Providing Hint:}
  The figure illustrates typical segmentation dimensions—\textit{personal}, \textit{behavioral}, \textit{geographical}, \textit{psychological}—that guide the user or the Marketing Manager in dissecting core consumer attributes. This \emph{hint provision} aligns with Section~3.4.2 in the main text, where BusiAgent injects domain knowledge or best practices (e.g., “consider location-based marketing or brand preference analysis”) to nudge the user or subordinate roles toward relevant solution pathways.

\item \textbf{Prompt Clarification:}
  The bottom panel addresses how BusiAgent zeroes in on the \emph{value proposition}—an essential part of the Business Model Canvas (see Figure 13) - explaining its positioning relative to competitors. For example, “Enterprises can attract customers by providing core values that differ from their competitors.” By systematically clarifying terms like “positioning” and “key value,” the system mitigates misunderstandings and ensures all roles share a unified interpretation of strategic priorities (e.g., Marketing Manager vs. CFO alignment on pricing and brand differentiation).

\end{itemize}

\noindent
\textbf{Connection to Contextual Thompson Sampling (Section~3.4).}\quad
Under the hood, BusiAgent treats each \textit{prompt refinement strategy} as a distinct 'arm' in a multiarmed bandit setting (Equation (2) in Section 3.4). The system then samples from a posterior distribution (contextualized by user and agent feedback), adjusting how it elaborates, provides hints, or clarifies the prompt. Over time, poorly performing prompt variants (e.g., insufficiently detailed elaborations) yield lower “reward” (less accurate role responses), while more effective prompts—like those that highlight user demographics or competitor strategies—gain higher reward. Thus, the \emph{chain-of-thought} approach (Section 3.3) systematically focuses on an optimal prompting style that addresses all relevant aspects (time constraints for CFO, brand concerns for CEO, data needs for CTO).

\noindent
\textbf{Avoiding Misalignment.}\quad
By performing multiple \textit{hint} or \textit{clarification} rounds, BusiAgent proactively circumvents incomplete or contradictory instructions. This is crucial for multi-agent synergy: the system ensures that each role's tasks are well-defined, and that roles requiring advanced data (e.g., CFO forecasting or CTO feasibility checks) receive explicit details upfront. In essence, Figure 4 exemplifies how BusiAgent's prompt optimization not only clarifies user requests but also aligns them with internal organizational roles---enhancing final outcomes and minimizing confusion within the enterprise decision pipeline.

\vspace{1em}
\noindent
\textbf{Figure 5 --- Quality Assurance Mechanism (LTM + STM + Knowledge Base)}\\
\textit{(See Section~3.5 for more details on QA.)}

Figure~5 illustrates how BusiAgent \emph{continuously maintains consistency} across extended multi-agent interactions by merging \textbf{Short-Term Memory (STM)}, \textbf{Long-Term Memory (LTM)}, and a \textbf{Knowledge Base}. This architecture underpins the system’s Quality Assurance (QA) process, preventing agents (CEO, CTO, CFO, etc.) from drifting into contradictory or suboptimal decisions when new information arrives.

\begin{itemize}[leftmargin=1.4em]
\item \textbf{Short-Term Memory (STM) and Long-Term Memory (LTM):}
  \begin{itemize}
  \item \emph{STM} caches the \emph{current conversation context} or \emph{partial solutions}, such as a newly proposed customer segment from the Marketing Manager or a just-updated budget figure from the CFO. It is the most recent “working set” of decisions and constraints that the agent must consider in real time.
  \item \emph{LTM} records historical constraints, actions taken in earlier sessions, and domain knowledge that has persisted across multiple interactions. For instance, the CFO’s budget cap decided \textbf{two weeks ago} or a marketing preference specified by the CEO \textbf{one month ago} remain accessible for cross-checking. 
  By referencing LTM, BusiAgent ensures that \emph{extended CTMDP policies} (Section~3.2) do not violate previously established constraints when a new round of decisions arises.
  \end{itemize}

\item \textbf{Knowledge Base (KB):}
  \begin{itemize}
  \item The KB holds curated \emph{domain references}, such as best practices, regulatory guidelines, or company policies. For example, it may contain industry-specific compliance rules for healthcare or finance, or a library of “known best practices” for marketing campaigns.
  \item When an agent’s action (e.g., “launch a global marketing strategy”) might conflict with domain rules (e.g., “data privacy restrictions”), the system queries the KB to catch potential issues before the plan is finalized.
  \end{itemize}

\item \textbf{Correction Mechanism:}
  \begin{enumerate}[label=(\alph*)]
  \item \emph{Conflict Detection}: The system checks newly generated outputs (or revised strategies) against STM, LTM, and the KB. For instance, if the PM proposes an expansion plan requiring more funds than the CFO’s budget allows, a mismatch is detected.
  \item \emph{Correction Trigger}: If conflicts arise—either internal (contradicting prior decisions) or external (violating domain guidelines)—BusiAgent triggers a correction phase. The relevant agent (e.g., the PM) is prompted to either adjust the proposal or seek clarification from the CFO or CEO.
  \item \emph{Resolution Loop}: The system repeats this verification cycle until no conflicts remain or a final user override occurs. This ensures the multi-agent environment remains faithful to previously set constraints, preventing contradictory solutions and enhancing overall coherence.
  \end{enumerate}
\end{itemize}

By continuously reconciling \emph{new statements} with \emph{historical context} (LTM) and \emph{domain expertise} (KB), BusiAgent’s QA mechanism (Section~3.5) preserves logical coherence across multiple steps or sessions. Figure~5 thus depicts a \textbf{self-correcting workflow}: each updated solution is validated, conflicts are surfaced, and roles are nudged to revise their plans if they contradict prior truths or best practices. This layered memory structure—STM for immediate context, LTM for persistent knowledge, and KB for domain constraints—enables BusiAgent to systematically avoid \emph{inconsistencies} and maintain a robust chain-of-thought in complex enterprise decision pipelines.

\vspace{1em}
\noindent
\textbf{Figure 6 --- Customer Segmentation Analysis}\\
\textit{(Discussed in Section~4.2 as a running example.)}

Figure~6 illustrates how a \emph{customer segmentation market report} request from the CEO flows through BusiAgent’s hierarchical multi-agent system. This scenario highlights the extended CTMDP approach (Section~3.2) and multi-level Stackelberg dynamics (Section~3.3), showing how each specialized role—CEO, CTO, Marketing Manager (MM), and Product Manager (PM)—collaborates to derive insights for an AI-based translation product.

\begin{itemize}[leftmargin=1.4em]
\item \textbf{CEO to CTO (Strategic Initiation):}  
  The CEO initiates a strategic question: “Which customer segments should we target for our AI translation solution?” 
  Under the multilevel Stackelberg framework (Theorem 2), the CEO directive guides subsequent decisions made by technical and operational roles.

\item \textbf{CTO to Marketing Manager (High-Level Delegation):}  
  The CTO, responsible for \emph{tech feasibility} and \emph{system integration}, delegates marketing-relevant inquiries to the MM. 
  This step reflects the hierarchical chain of command: roles further down the hierarchy accept tasks from higher-level roles, 
  consistent with Definition 3 of the multi-level game. 
  In practice, the MM might gather user profiles or examine prior marketing data from the CFO’s budgeting constraints or historical spending.

\item \textbf{Marketing Manager to Product Manager (Detailed Analysis):}  
  Once the MM defines the scope (“We need data-driven segmentation”), the PM is tasked with executing advanced analytics. 
  As indicated in the figure, the PM can invoke the \texttt{Python\_Executor()} (Section 3.4) to run clustering (e.g., PCA, K-means).
  This chain-of-thought step connects back to the \emph{Action} space in the extended CTMDP: 
  an agent selects a specialized tool (Equation (1)) to reduce uncertainty or meet new sub-goals (like identifying clusters of enterprise vs.\ individual consumers).

\item \textbf{PM Performs Data Analytics and Reports Back (Execution + Feedback):}  
  The bottom-right portion of Figure 6 shows outputs like PCA-based visualizations and morphological analyses of user segments
  (e.g., corporations, government agencies, translation and localization specialists). 
  After obtaining these analytical insights, the PM reports results back up the chain—to the CTO—ensuring that final strategic decisions (CEO-level) 
  reflect robust, data-driven segmentation solutions. 
  This "reporting work" loop (Section 3.3) maintains alignment between lower-level analysis and higher-level strategy.

\end{itemize}

\noindent
\textbf{Synergy of Hierarchical Roles and Tools.}\quad
By depicting \textbf{task delegation} (CEO $\to$ CTO $\to$ MM $\to$ PM) and \textbf{reporting work} (PM $\to$ CTO $\to$ CEO), 
Figure 6 exemplifies how each role focuses on tasks within its domain while collaboratively tackling a real business question. 
The CEO sets overarching objectives, the CTO ensures technical feasibility, the Marketing Manager frames the market context, and the Product Manager executes detailed data analysis. 
Within BusiAgent, each agent’s extended CTMDP policy captures both time-sensitive decisions ($\omega_i$) and the multi-level Stackelberg structure, 
so that the entire pipeline—\emph{from initial strategic prompt to final segmented insights}—operates cohesively.

\noindent
\textbf{Practical Takeaway.}\quad
This example underscores BusiAgent’s ability to resolve complex, cross-departmental tasks (e.g., bridging marketing strategy, technical analytics, and executive decision-making). 
It also illustrates how specialized tools (e.g., Python-based analytics) integrate seamlessly into the role-based workflow, 
transforming raw data into actionable knowledge for top-level decisions. 
Ultimately, Figure 6 showcases the hallmark of the framework: 
uniting “bits” (data clustering, morphological analysis) with “boardrooms” (CEO’s segmentation strategy) in a single multi-agent environment.

\vspace{1em}
\noindent
\textbf{Concluding Remarks on Figures 2–6:}\\
These six figures represent the full breadth of BusiAgent’s operation:
\begin{itemize}[leftmargin=1.4em]
\item \textbf{Figure 2} gives the broad system pipeline with user prompts, multi-role agents, tool usage, and chain-of-thought.  
\item \textbf{Figure 3} enumerates the specific tools each agent can invoke in its extended CTMDP action space.  
\item \textbf{Figure 4} shows a typical delegation chain for a customer-segmentation use case, reflecting multi-level Stackelberg collaboration.  
\item \textbf{Figure 5} highlights how prompt refinement (Thompson sampling) leads to more complete or focused solutions.  
\item \textbf{Figure 6} demonstrates short-/long-term memory plus a knowledge base working together to preserve consistency and correctness.  
% \item \textbf{Figure 7} depicts a full business model canvas example, unifying the roles, tools, and QA processes into a practical scenario for an AI translation startup.  
\end{itemize}

\bigskip

\section*{Appendix G: Business Model Canvas and Role-based Workflow}
\label{appendix:simulation_ops}

\begin{figure*}[t]
\centering
\includegraphics[width=\linewidth]{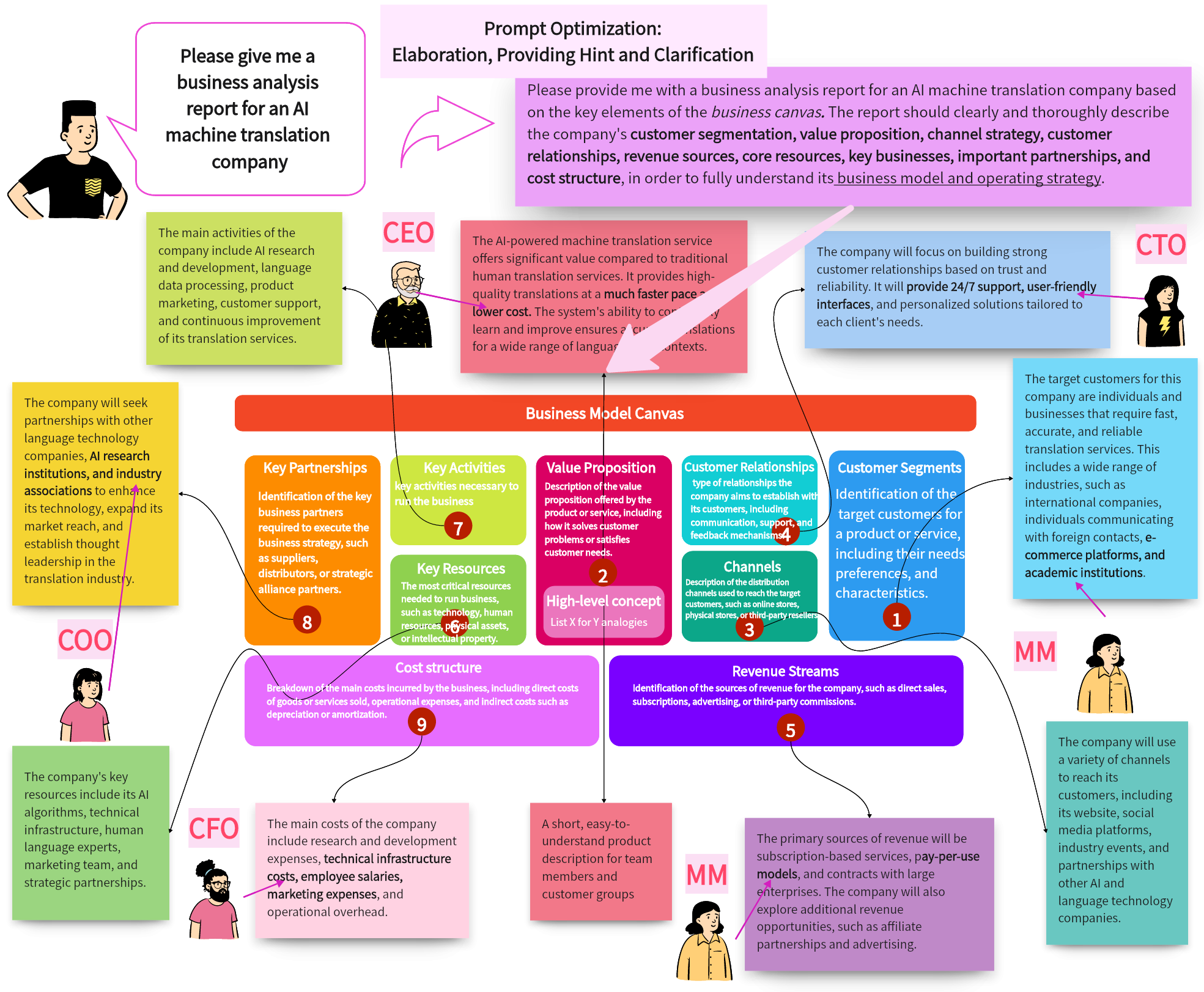}
\caption{Business Model Canvas and Role-based Workflow: translation startup demonstration.}
\label{fig:simulation-ops}
\end{figure*}

\noindent
\textbf{Overview.}
Figure~\ref{fig:simulation-ops} illustrates how \textbf{BusiAgent} applies a \emph{Business Model Canvas} (BMC) approach in conjunction with hierarchical role-based workflows to analyze a machine translation (MT) startup. The canvas is organized into nine standard blocks (customer segments, value proposition, channels, customer relationships, revenue streams, key resources, key activities, key partnerships, and cost structure), which together capture the essential components of a technology-based business model.

\paragraph{BMC Blocks (Center).}
In the middle of Figure~\ref{fig:simulation-ops}, colored rectangles outline each BMC building block:
\begin{enumerate}[label=(\arabic*)]
\item \textbf{Customer Segments} — The target audiences for a machine translation product, such as e-commerce platforms, multinational companies, and academic institutions needing fast, accurate translations.
\item \textbf{Value Proposition} — Explains how the MT service delivers unique benefits (e.g., faster turnaround, cost savings, or multilingual support) compared to traditional translation options.
\item \textbf{Channels} — The routes used to reach customers (websites, partner alliances, industry events).
\item \textbf{Customer Relationships} — Approaches to maintain or grow customer loyalty (24/7 support, personalization, user-friendly interfaces).
\item \textbf{Revenue Streams} — Monetization models (subscription-based services, pay-per-use options, third-party commissions).
\item \textbf{Key Activities} — Core operations needed to run the service (AI model R\&D, language data processing, platform maintenance).
\item \textbf{Key Resources} — Vital assets (infrastructure, skilled personnel, proprietary code or algorithms).
\item \textbf{Key Partnerships} — Collaborations (e.g., alliances with other language-tech firms or AI research institutions) that enhance technology or market reach.
\item \textbf{Cost Structure} — Cost breakdown (R\&D, salaries, technical infrastructure, marketing).
\end{enumerate}

\paragraph{Roles and Workflows (Surrounding Boxes).}
Different roles (CEO, CTO, COO, CFO, Marketing Manager) appear around the canvas, each focusing on one or more BMC blocks according to their CTMDP-driven responsibilities (Sections~3.2–3.3 of the main text):
\begin{itemize}
\item \textbf{CEO} (top center) focuses on overall strategy and leadership, coordinating how the BMC blocks fit into a coherent business vision.
\item \textbf{CTO} (right side) tackles the technical feasibility of each block’s requirements, e.g., ensuring the infrastructure can handle multilingual throughput and real-time user demands.
\item \textbf{COO} (left side) explores potential alliances with other language-technology companies or research labs, aligning with “Key Partnerships” and “Key Activities.”
\item \textbf{CFO} (bottom left) weighs in on cost structure, controlling budgets for R\&D, marketing, and salaries.
\item \textbf{Marketing Manager (MM)} (bottom right) refines “Customer Segments,” “Channels,” and “Customer Relationships,” developing strategies to expand usage scenarios and brand presence.
\end{itemize}

\paragraph{Prompt Optimization and QA Mechanisms.}
At the top, a prompt referencing “Please give me a business analysis report...” showcases how BusiAgent transforms a single, high-level user query into multiple sub-requests (customer segmentation, value proposition, revenue streams, etc.). The figure also indicates how BusiAgent’s \emph{Prompt Optimization} (Section~3.4) and \emph{Quality Assurance} (Section~3.5) come into play: each role’s partial contributions are refined through iterative prompts and validated for consistency against the knowledge base and short-/long-term memory (Figures~4--5 in Appendix F).

\paragraph{Integration with Multi-Agent Workflow.}
Arrows link roles (e.g., CEO $\rightarrow$ CTO, CFO, COO), illustrating how tasks pass down or horizontally between departments. This corresponds to the multi-level Stackelberg game (Theorem 3) in Section~3.3 and the extended CTMDP approach in Section~3.2. For instance:
\begin{itemize}
\item The \textbf{CEO} gives strategic directives (e.g., “focus on e-commerce platforms”).
\item The \textbf{CTO} coordinates the technical tasks with the Product Manager or data scientists.
\item The \textbf{COO} manages external partnerships, ensuring alliances align with cost structure constraints from the \textbf{CFO}.
\end{itemize}

\paragraph{Significance.}
By mapping the BMC to role-specific responsibilities, BusiAgent ensures that strategic directions (CEO-level) and operational details (manager-level) remain synchronized. The resulting synergy eliminates organizational silos, accelerates response times, and leads to more robust decisions—points validated in the experimental evaluations (Section~4.2). Figure~\ref{fig:simulation-ops} thus encapsulates the \emph{“from bits to boardrooms”} journey that underpins BusiAgent’s design philosophy, concretely demonstrating how \textbf{LLM-driven multi-agent workflows} can streamline complex business analyses in real-world contexts.

\bigskip
\noindent
%======================================================================
%  Appendix H  —  Pseudocode for BusiAgent Orchestration (English only)
%======================================================================
\section*{Appendix H: Pseudocode and Detailed Explanation}
\label{appendix:pseudocode}

%% ----------------------------------------------------------
%% 2.  Full-width algorithm*  (use algorithm if single-column)
%% ----------------------------------------------------------
\begin{algorithm*}[t]
\small          %% switch to \footnotesize if you still overflow
\caption{High-Level Orchestration of \textsc{BusiAgent}}
\label{alg:master}
\begin{algorithmic}[1]

%% ---------------- 2.1  Input / Output ---------------------
\Require
  $\{(S_i,A_i,q_i,r_i,\gamma_i,\omega_i)\}_{i=1..N}$: extended CTMDP of each role\\
  Stackelberg levels $L=\{1,2,\dots,L_{\max}\}$ (one level per agent)\\
  shared memory $\mathcal{M}$,\; knowledge $\mathcal{K}$,\; tool set $\mathcal{T}$\\
  initial policy set $\Pi=\{\pi_i\}_{i=1..N}$
\Ensure  coherent multi-role plan or updated policies

%% ---------------- 2.2  Main function ----------------------
\Function{BusiAgent-Orchestrate}{}
  \State initialise $\mathcal{M}\gets\varnothing$;\; load $\mathcal{K}$;\; reset all $s_i\in S_i$
  \State $t\gets0$

  \While{\textit{not converged}}
    %% ---- Vertical coordination (Stackelberg) ------------
    \Statex\(\triangleright\) \textit{Vertical coordination: multi-level Stackelberg}
    \ForAll{levels $l\in L$ (top-down)}
      \ForAll{agents $i$ on level $l$}
        \State observe $s_i(t)$
        \State choose
        \[
          a_i(t)=\arg\max_{a\in A_i}\Bigl[
            r_i(s_i,a)
            +\!\!\int_{0}^{\omega_i(s_i,a)}\!\!
              e^{-\gamma_i \tau}\!
              \sum_{s'\in S_i}q_i(s_i,a,s')V_i^{*}(s')\,d\tau
          \Bigr]
        \]
        \If{$a_i(t)$ calls a tool from $\mathcal{T}$}
          \State execute the requested operation
        \EndIf
        \State update STM;\; push key facts to global $\mathcal{M}$
      \EndFor
      \Statex\(\triangleright\) level $l{+}1$ agents now read updated states
    \EndFor

    %% ---- Horizontal collaboration (brainstorming) -------
    \Statex\(\triangleright\) \textit{Horizontal collaboration: brainstorming}
    \State gather partial solutions;\; compute posterior $p(x\mid y)$ and entropy $H_\alpha$
    \State refine proposals if $D_\alpha(p\|q)\ge\epsilon$

    %% ---- QA check ---------------------------------------
    \Statex\(\triangleright\) \textit{Quality assurance}
    \State cross-check with LTM and $\mathcal{K}$
    \If{conflict detected}
      \State trigger correction loop until resolved
    \EndIf

    %% ---- Prompt optimisation (Thompson) -----------------
    \Statex\(\triangleright\) \textit{Prompt optimisation (Thompson sampling)}
    \State observe context $x_t$
    \ForAll{prompt variants $k$}
      \State sample $\theta_k(t)$;\; pick $k^{*}=\arg\max\theta_k(t)(x_t)$
      \State execute $k^{*}$;\; receive reward $r_t\in[0,1]$;\; update GP posterior
    \EndFor

    \State check convergence;\; $t\gets t+1$
  \EndWhile
  \State \Return final plan or $\{\pi_i^{*}\}$
\EndFunction
\end{algorithmic}
\end{algorithm*}

\noindent
\textbf{Detailed Explanation of Pseudocode in Algorithm~\ref{alg:master}:}\\[4pt]

\textbf{1--8.\ Initialization}\\[-2pt]
\begin{itemize}[leftmargin=1.5em]
  \item We collect extended CTMDP parameters
        $\{(S_i,A_i,q_i,r_i,\gamma_i,\omega_i)\}$ for each role~$i$.
  \item The system starts with empty (or lightly seeded) memory
        $\mathcal{M}$—combining short/long-term memory from
        Section~3.5—and a curated knowledge base~$\mathcal{K}$.
  \item Every role’s local state~$s_i$ is initialised (e.g., the CFO’s
        state may contain budget constraints, while the CTO’s state
        includes project-feasibility data).
  \item Policies $\pi_i$ can be default or pretrained; either way they
        will be refined inside the main loop.
\end{itemize}

\textbf{9--25.\ Main Orchestration While-Loop}\\[-2pt]
\begin{itemize}[leftmargin=1.5em]
  \item The system iterates until it \emph{converges}: all roles hold
        stable plans or the user signals satisfaction (stopping
        criterion).
\end{itemize}

\textbf{Lines 10--17: Vertical Coordination (Multi-level Stackelberg)}\\[-2pt]
\begin{itemize}[leftmargin=1.5em]
  \item We iterate from the highest level $l=1$ (e.g., CEO) downward to
        lower levels $l=2,3,\dots$.
  \item For each agent $i$ at level $l$:
        \begin{enumerate}[label=\arabic*.,leftmargin=1.5em]
          \item The agent observes its local state~$s_i(t)$, including
                the updated memory~$\mathcal{M}$.
          \item It selects an action $a_i(t)\in A_i$ by maximising
                the integrated reward plus future value over
                $\omega_i$ time (Theorem~1 / Eq.~(1)).
          \item If the action involves a specialised tool call
                (e.g., \texttt{Python\_Executor} or
                \texttt{DuckDuckGo\_Search}), the agent executes it and
                retrieves new data.
          \item Finally, it updates short-term memory and may push key
                information to the global memory~$\mathcal{M}$.
        \end{enumerate}
  \item This top-down pass enforces Stackelberg logic (Theorem~2): roles
        at lower levels must adapt to decisions made above.
\end{itemize}

\textbf{Lines 19--22: Horizontal Collaboration (Entropy-based Brainstorming)}\\[-2pt]
\begin{itemize}[leftmargin=1.5em]
  \item After each vertical pass, roles on the same (or neighbouring)
        levels exchange partial solutions.
  \item The system computes the posterior $p(x\mid y)$ and checks the
        generalised entropy $H_{\alpha}(X)$.  
        If the Rényi divergence $D_{\alpha}\ge\epsilon$, faster
        consensus is expected (Theorem~2).
  \item Partial proposals are merged or refined for maximal synergy
        before the QA phase.
\end{itemize}

\textbf{Lines 24--29: QA Checking and Correction Mechanism}\\[-2pt]
\begin{itemize}[leftmargin=1.5em]
  \item New outputs are cross-checked against LTM, domain constraints in
        $\mathcal{K}$, and existing budgets/policies.
  \item If contradictions arise (e.g., CFO cost structure exceeded or
        compliance rules violated), a correction loop is triggered until
        no conflict remains.
\end{itemize}

\textbf{Lines 31--39: Prompt Optimisation / Thompson Sampling}\\[-2pt]
\begin{itemize}[leftmargin=1.5em]
  \item We incorporate bandit-based prompt optimisation
        (Section~3.4). The system observes context~$x_t$
        (e.g., task complexity or domain) and tests
        different prompt variants~$k$.
  \item Each variant’s performance is scored by a reward
        $r_t\in[0,1]$. Over multiple iterations the Gaussian-process
        posteriors $(\mu_k,K_k)$ are updated, ensuring prompts align
        better with each role’s needs.
\end{itemize}

\textbf{Lines 41--42: Convergence and Return}\\[-2pt]
\begin{itemize}[leftmargin=1.5em]
  \item After every iteration, the system checks for a stopping
        condition (stable policies, user acceptance, or lack of further
        improvement).
  \item Once converged, \textsc{BusiAgent} outputs the multi-agent plan
        or the updated policy set $\{\pi_i^{*}\}$.
\end{itemize}

\medskip
\noindent
\textbf{Key Takeaways}\\[-4pt]
\begin{itemize}[leftmargin=1.5em]
  \item \emph{Lines 10 ff.} show how high-level decisions cascade
        downward in Stackelberg fashion.
  \item \emph{Lines 19–22} illustrate horizontal brainstorming, unifying
        partial solutions via entropy metrics.
  \item \emph{Lines 24–29} highlight the QA layer (STM/LTM and knowledge
        base) that enforces consistency.
  \item \emph{Lines 31–39} detail how Thompson sampling tunes prompt
        variants for better role synergy.
\end{itemize}

Overall, Algorithm~\ref{alg:master} fuses CTMDP control, multi-level
Stackelberg coordination, entropy-driven brainstorming, rigorous QA, and
bandit-style prompt optimisation into one cohesive orchestration loop,
realising the “bits to boardrooms” workflow described in the main text.

\noindent
\textbf{End of Appendices.} \\
We have comprehensively updated and enriched the appendices (A--G) to ensure they effectively complement the main text:

\begin{itemize}[leftmargin=1.3em]
    \item \textbf{Unique numbering for all formulas and theorems.}  
    Each result in the main text now has a corresponding extended proof in Appendix~C, with carefully synchronized labels. This ensures that every theorem, equation, or lemma in the main text can be directly traced to its detailed mathematical background, eliminating ambiguity and streamlining cross-referencing.

    \item \textbf{Enhanced content in Appendices A--E.}  
    In these sections, we provide an extensive array of code snippets (e.g., DuckDuckGo Search, HuggingFace Tools), elaborated chain-of-thought transcripts for various roles (CEO, CFO, etc.), and deeper explorations of domain-specific scenarios (e.g., education, healthcare, finance). Together, these additions illuminate how the extended CTMDP, multi-level Stackelberg modeling, and contextual Thompson sampling are practically instantiated within BusiAgent’s architecture.

    \item \textbf{Figure-by-figure commentaries in Appendix~F.}  
    We offer thorough explanations for each key figure (Figures~2--6), clarifying how they depict the core components of BusiAgent, including CTMDP-based decision-making, hierarchical role delegation, iterative brainstorming, and prompt optimization. By connecting each visual to specific sections or theorems in the main paper, readers can see precisely how conceptual designs (like multi-level Stackelberg coordination) map onto real-world workflow diagrams.

    \item \textbf{New Appendix~G with a Business Model Canvas demonstration.}  
    Lastly, we introduce a new appendix dedicated to illustrating how BusiAgent applies role-based workflows and prompt optimization to develop a comprehensive Business Model Canvas for a machine translation startup. This final figure (Figure~\ref{fig:simulation-ops}) exemplifies the system’s capacity to unify strategic directives (CEO), resource allocation (CFO, COO), technical constraints (CTO), and marketing insights (Marketing Manager) into a cohesive business plan.
\end{itemize}

Taken together, these expanded materials provide a more holistic, in-depth view of \textbf{BusiAgent}’s design, theoretical underpinnings, and real-world applicability, ensuring that readers have ample resources to fully grasp the framework’s capabilities and implementation details.

%%%%%%%%%%%%%%%%%%%%%%%%%%%%%%%%%%%%%%%%%%%%%%%%%%%%%%%%%%%%%%%%%%%%%%%%

%%% Use this command to include your bibliography file.
\clearpage
\bibliography{mybibfile}

@misc{mckinsey2025ai,
  title        = {The State of Generative AI in the Enterprise 2025},
  author       = {McKinsey \& Company},
  howpublished = {McKinsey Global Institute Report},
  year         = {2025},
  month        = apr,
  note         = {Section 3, p.~18 reports a projected 1.5\,pp annual productivity uplift from multi-agent AI adoption},
  url          = {https://www.mckinsey.com/business-functions/quantumblack/our-insights/state-of-ai-2025}
}

@misc{autogenstudio2024,
  title        = {AutoGen Studio: A No-Code Developer Tool for Multi-Agent Workflows},
  author       = {Ruiqi Jiang and Zihan Wang and Shuyue Hu and Jie Fu and Xiaoyong Jin},
  howpublished = {arXiv preprint arXiv:2408.15247},
  year         = {2024},
  month        = aug,
  url          = {https://arxiv.org/abs/2408.15247}
}

@misc{gpt4o2024,
  title        = {{GPT}\hyp{}4o System Card},
  author       = {{OpenAI}},
  howpublished = {OpenAI Technical Report},
  year         = {2024},
  month        = may,
  url          = {https://cdn.openai.com/papers/gpt4o-system-card.pdf}
}

@misc{gemini15,
  title        = {Gemini~1.5 Technical Report},
  author       = {{Google DeepMind}},
  howpublished = {DeepMind Research Report},
  year         = {2025},
  month        = feb,
  url          = {https://storage.googleapis.com/deepmind-media/gemini/gemini_15_report.pdf}
}

@misc{claude3,
  title        = {Claude~3 Model Card},
  author       = {{Anthropic}},
  howpublished = {Anthropic Documentation},
  year         = {2024},
  month        = mar,
  url          = {https://www.anthropic.com/research/claude-3-model-card}
}

@misc{llmagentSurvey2025,
  title        = {LLM Agents in 2025: A Comprehensive Survey and Roadmap},
  author       = {Chenxi Liu and Xiang Ren and Diyi Yang},
  howpublished = {arXiv preprint arXiv:2504.01234},
  year         = {2025},
  url          = {https://arxiv.org/abs/2504.01234}
}

@misc{autoagent2025,
  title        = {AutoAgent: A Fully\hyp{}Automated and Zero\hyp{}Code Framework for LLM Agents},
  author       = {Jiabin Tang and Tianyu Fan and Chao Huang},
  howpublished = {arXiv preprint arXiv:2502.05957},
  year         = {2025},
  url          = {https://arxiv.org/abs/2502.05957}
}

@misc{agentverse2023,
  title        = {AgentVerse: Facilitating the Development of Multi\hyp{}Agent LLM Systems},
  author       = {Tong Zhou and Yida Liu and Kun Shao},
  howpublished = {arXiv preprint arXiv:2310.01844},
  year         = {2023},
  url          = {https://arxiv.org/abs/2310.01844}
}

@misc{xagent2024,
  title        = {XAgent: A Versatile LLM Agent for Embodied Reasoning},
  author       = {Zhiying Jiang and Wei Shen and Chen Henry Wu},
  howpublished = {arXiv preprint arXiv:2403.06789},
  year         = {2024},
  url          = {https://arxiv.org/abs/2403.06789}
}

@misc{masllm2025,
  title        = {Multi\hyp{}Agent Systems Powered by Large Language Models: Applications in Swarm Intelligence},
  author       = {Cristian Jimenez\hyp{}Romero and Alper Yegenoglu and Christian Blum},
  howpublished = {arXiv preprint arXiv:2503.03800},
  year         = {2025},
  url          = {https://arxiv.org/abs/2503.03800}
}

@misc{multiagentbench2025,
  title        = {MultiAgentBench: Benchmarking Collaborative Reasoning of LLM\hyp{}Based Agents},
  author       = {Haonan Lu and Zeqiu Wu and Ziming Ding},
  howpublished = {arXiv preprint arXiv:2502.08888},
  year         = {2025},
  url          = {https://arxiv.org/abs/2502.08888}
}

@misc{sirius2025,
  title        = {SiriuS: Self\hyp{}Improving Multi\hyp{}Agent Systems via Continual Skill Discovery},
  author       = {Qiyuan Zhang and Rui Meng and Nenad Tomasev},
  howpublished = {arXiv preprint arXiv:2503.05555},
  year         = {2025},
  url          = {https://arxiv.org/abs/2503.05555}
}

@misc{talkhier2025,
  title        = {TalkHier: Hierarchical Communication Protocols for Large Language Model Agents},
  author       = {Haojie Li and Yichong Xu and Jie Fu},
  howpublished = {arXiv preprint arXiv:2501.09999},
  year         = {2025},
  url          = {https://arxiv.org/abs/2501.09999}
}

@article{tang2025autoagent,
  title={AutoAgent: A Fully-Automated and Zero-Code Framework for LLM Agents},
  author={Tang, Jiabin and Fan, Tianyu and Huang, Chao},
  journal={arXiv e-prints},
  pages={arXiv--2502},
  year={2025}
}

@article{jimenez2025multi,
  title={Multi-Agent Systems Powered by Large Language Models: Applications in Swarm Intelligence},
  author={Jimenez-Romero, Cristian and Yegenoglu, Alper and Blum, Christian},
  journal={arXiv preprint arXiv:2503.03800},
  year={2025}
}

@article{li2023metaagents,
  title={Metaagents: Simulating interactions of human behaviors for llm-based task-oriented coordination via collaborative generative agents},
  author={Li, Yuan and Zhang, Yixuan and Sun, Lichao},
  journal={arXiv preprint arXiv:2310.06500},
  year={2023}
}

@article{ouyang2022training,
  title={Training language models to follow instructions with human feedback},
  author={Ouyang, Long and Wu, Jeffrey and Jiang, Xu and Almeida, Diogo and Wainwright, Carroll and Mishkin, Pamela and Zhang, Chong and Agarwal, Sandhini and Slama, Katarina and Ray, Alex and others},
  journal={Advances in neural information processing systems},
  volume={35},
  pages={27730--27744},
  year={2022}
}

@article{openai2023gpt4,
  title={Gpt-4 technical report},
  author={Achiam, Josh and Adler, Steven and Agarwal, Sandhini and Ahmad, Lama and Akkaya, Ilge and Aleman, Florencia Leoni and Almeida, Diogo and Altenschmidt, Janko and Altman, Sam and Anadkat, Shyamal and others},
  journal={arXiv preprint arXiv:2303.08774},
  year={2023}
}

@article{wei2022emergent,
  title={Emergent abilities of large language models},
  author={Wei, Jason and Tay, Yi and Bommasani, Rishi and Raffel, Colin and Zoph, Barret and Borgeaud, Sebastian and Yogatama, Dani and Bosma, Maarten and Zhou, Denny and Metzler, Donald and others},
  journal={arXiv preprint arXiv:2206.07682},
  year={2022}
}

@article{brown2020language,
  title={Language models are few-shot learners},
  author={Brown, Tom and Mann, Benjamin and Ryder, Nick and Subbiah, Melanie and Kaplan, Jared D and Dhariwal, Prafulla and Neelakantan, Arvind and Shyam, Pranav and Sastry, Girish and Askell, Amanda and others},
  journal={Advances in neural information processing systems},
  volume={33},
  pages={1877--1901},
  year={2020}
}

@inproceedings{park2023generative,
  title={Generative agents: Interactive simulacra of human behavior},
  author={Park, Joon Sung and O'Brien, Joseph and Cai, Carrie Jun and Morris, Meredith Ringel and Liang, Percy and Bernstein, Michael S},
  booktitle={Proceedings of the 36th annual acm symposium on user interface software and technology},
  pages={1--22},
  year={2023}
}

@article{hong2023metagpt,
  title={Metagpt: Meta programming for multi-agent collaborative framework},
  author={Hong, Sirui and Zheng, Xiawu and Chen, Jonathan and Cheng, Yuheng and Wang, Jinlin and Zhang, Ceyao and Wang, Zili and Yau, Steven Ka Shing and Lin, Zijuan and Zhou, Liyang and others},
  journal={arXiv preprint arXiv:2308.00352},
  year={2023}
}

@article{zhuge2023mindstorms,
  title={Mindstorms in natural language-based societies of mind},
  author={Zhuge, Mingchen and Liu, Haozhe and Faccio, Francesco and Ashley, Dylan R and Csord{\'a}s, R{\'o}bert and Gopalakrishnan, Anand and Hamdi, Abdullah and Hammoud, Hasan Abed Al Kader and Herrmann, Vincent and Irie, Kazuki and others},
  journal={arXiv preprint arXiv:2305.17066},
  year={2023}
}

@article{cai2023large,
  title={Large language models as tool makers},
  author={Cai, Tianle and Wang, Xuezhi and Ma, Tengyu and Chen, Xinyun and Zhou, Denny},
  journal={arXiv preprint arXiv:2305.17126},
  year={2023}
}

@misc{wang2023unleashing,
      title={Unleashing Cognitive Synergy in Large Language Models: A Task-Solving Agent through Multi-Persona Self-Collaboration}, 
      author={Zhenhailong Wang and Shaoguang Mao and Wenshan Wu and Tao Ge and Furu Wei and Heng Ji},
      year={2023},
      eprint={2307.05300},
      archivePrefix={arXiv},
      primaryClass={cs.AI}
}

@article{nakano2021webgpt,
  title={Webgpt: Browser-assisted question-answering with human feedback},
  author={Nakano, Reiichiro and Hilton, Jacob and Balaji, Suchir and Wu, Jeff and Ouyang, Long and Kim, Christina and Hesse, Christopher and Jain, Shantanu and Kosaraju, Vineet and Saunders, William and others},
  journal={arXiv preprint arXiv:2112.09332},
  year={2021}
}

@article{yao2022react,
  title={React: Synergizing reasoning and acting in language models},
  author={Yao, Shunyu and Zhao, Jeffrey and Yu, Dian and Du, Nan and Shafran, Izhak and Narasimhan, Karthik and Cao, Yuan},
  journal={arXiv preprint arXiv:2210.03629},
  year={2022}
}

@article{schick2023toolformer,
  title={Toolformer: Language models can teach themselves to use tools},
  author={Schick, Timo and Dwivedi-Yu, Jane and Dess{\`\i}, Roberto and Raileanu, Roberta and Lomeli, Maria and Hambro, Eric and Zettlemoyer, Luke and Cancedda, Nicola and Scialom, Thomas},
  journal={Advances in Neural Information Processing Systems},
  volume={36},
  pages={68539--68551},
  year={2023}
}

@article{lu2024chameleon,
  title={Chameleon: Plug-and-play compositional reasoning with large language models},
  author={Lu, Pan and Peng, Baolin and Cheng, Hao and Galley, Michel and Chang, Kai-Wei and Wu, Ying Nian and Zhu, Song-Chun and Gao, Jianfeng},
  journal={Advances in Neural Information Processing Systems},
  volume={36},
  year={2024}
}

@misc{qin2023tool,
      title={Tool Learning with Foundation Models}, 
      author={Yujia Qin and Shengding Hu and Yankai Lin and Weize Chen and Ning Ding and Ganqu Cui and Zheni Zeng and Yufei Huang and Chaojun Xiao and Chi Han and Yi Ren Fung and Yusheng Su and Huadong Wang and Cheng Qian and Runchu Tian and Kunlun Zhu and Shihao Liang and Xingyu Shen and Bokai Xu and Zhen Zhang and Yining Ye and Bowen Li and Ziwei Tang and Jing Yi and Yuzhang Zhu and Zhenning Dai and Lan Yan and Xin Cong and Yaxi Lu and Weilin Zhao and Yuxiang Huang and Junxi Yan and Xu Han and Xian Sun and Dahai Li and Jason Phang and Cheng Yang and Tongshuang Wu and Heng Ji and Zhiyuan Liu and Maosong Sun},
      year={2023},
      eprint={2304.08354},
      archivePrefix={arXiv},
      primaryClass={cs.CL}
}

@software{Significant_Gravitas_AutoGPT,
    author = {{Significant Gravitas}},
    license = {MIT},
    title = {{AutoGPT}},
    url = {https://github.com/Significant-Gravitas/AutoGPT},
    year = {2023}
}

@software{WorkGPT,
    author = {{Team OPENPM}},
    license = {MIT},
    title = {{WorkGPT}},
    url = {https://github.com/team-openpm/workgpt},
    year = {2023}
}

@software{gpt-engineer,
    author = {{Osika Anton}},
    license = {MIT},
    title = {{gpt-engineer}},
    url = {https://github.com/AntonOsika/gpt-engineer},
    year = {2023}
}

@software{SmolModels,
    author = {{SMOL AI}},
    license = {MIT},
    title = {{SmolModels}},
    url = {https://github.com/smol-ai/developer},
    year = {2023}
}

@misc{openai_2023_CI,
  title        = {ChatGPT plugins},
  author       = {OpenAI},
  year         = 2023,
  month        = {March},
  day          = {23},
  howpublished = {\url{https://openai.com/blog/chatgpt-plugins}},
  note         = {Accessed: November 21, 2023}
}

@misc{langchain_2023_agents,
  title        = {Langchain Agents},
  author       = {LangChain, Inc.},
  year         = 2023,
  howpublished = {\url{https://python.langchain.com/docs/modules/agents/}},
  note         = {Accessed: November 21, 2023}
}

@article{du2023improving,
  title={Improving factuality and reasoning in language models through multiagent debate},
  author={Du, Yilun and Li, Shuang and Torralba, Antonio and Tenenbaum, Joshua B and Mordatch, Igor},
  journal={arXiv preprint arXiv:2305.14325},
  year={2023}
}

@article{liang2023encouraging,
  title={Encouraging divergent thinking in large language models through multi-agent debate},
  author={Liang, Tian and He, Zhiwei and Jiao, Wenxiang and Wang, Xing and Wang, Yan and Wang, Rui and Yang, Yujiu and Shi, Shuming and Tu, Zhaopeng},
  journal={arXiv preprint arXiv:2305.19118},
  year={2023}
}

@article{li2023camel,
  title={Camel: Communicative agents for" mind" exploration of large language model society},
  author={Li, Guohao and Hammoud, Hasan and Itani, Hani and Khizbullin, Dmitrii and Ghanem, Bernard},
  journal={Advances in Neural Information Processing Systems},
  volume={36},
  pages={51991--52008},
  year={2023}
}

@article{qian2023communicative,
  title={Communicative agents for software development},
  author={Qian, Chen and Cong, Xin and Yang, Cheng and Chen, Weize and Su, Yusheng and Xu, Juyuan and Liu, Zhiyuan and Sun, Maosong},
  journal={arXiv preprint arXiv:2307.07924},
  volume={6},
  number={3},
  year={2023}
}

@article{wu2023autogen,
  title={Autogen: Enabling next-gen llm applications via multi-agent conversation framework},
  author={Wu, Qingyun and Bansal, Gagan and Zhang, Jieyu and Wu, Yiran and Zhang, Shaokun and Zhu, Erkang and Li, Beibin and Jiang, Li and Zhang, Xiaoyun and Wang, Chi},
  journal={arXiv preprint arXiv:2308.08155},
  year={2023}
}

@article{rossetto2021conservation,
  title={A conservation genomics workflow to guide practical management actions},
  author={Rossetto, Maurizio and Yap, Jia-Yee Samantha and Lemmon, Jedda and Bain, David and Bragg, Jason and Hogbin, Patricia and Gallagher, Rachael and Rutherford, Susan and Summerell, Brett and Wilson, Trevor C},
  journal={Global ecology and conservation},
  volume={26},
  pages={e01492},
  year={2021},
  publisher={Elsevier}
}

@article{harper2015movielens,
  title={The movielens datasets: History and context},
  author={Harper, F Maxwell and Konstan, Joseph A},
  journal={Acm transactions on interactive intelligent systems (tiis)},
  volume={5},
  number={4},
  pages={1--19},
  year={2015},
  publisher={Acm New York, NY, USA}
}

@article{csiszar1995generalized,
  title={Generalized cutoff rates and R{\'e}nyi's information measures},
  author={Csisz{\'a}r, Imre},
  journal={IEEE Transactions on information theory},
  volume={41},
  number={1},
  pages={26--34},
  year={1995},
  publisher={IEEE}
}

@article{srinivas2009gaussian,
  title={Gaussian process optimization in the bandit setting: No regret and experimental design},
  author={Srinivas, Niranjan and Krause, Andreas and Kakade, Sham M and Seeger, Matthias},
  journal={arXiv preprint arXiv:0912.3995},
  year={2009}
}

\end{document}